\documentclass[10pt,twocolumn,letterpaper]{article}

\usepackage[pagenumbers]{cvpr} %

\usepackage[accsupp]{axessibility}  %

\usepackage{url}            %
\usepackage{booktabs}       %
\usepackage{amsfonts}       %
\usepackage{nicefrac}       %
\usepackage{microtype}      %
\usepackage{xcolor}         %
\usepackage{multirow}
\usepackage{graphicx}
\usepackage{enumitem} %
\usepackage{amsmath}

\usepackage{caption} %
\usepackage{float} %
\usepackage{bm}

\usepackage{subcaption}
\usepackage{tabularx}
\usepackage{makecell}
\usepackage{pifont}

\newcommand{\boldstartspace}[1]{\vspace{0.2em}\noindent\textbf{#1}}

\newcommand{\new}[1]{{#1}}

\definecolor{cvprblue}{rgb}{0.21,0.49,0.74}
\usepackage[pagebackref,breaklinks,colorlinks,allcolors=cvprblue]{hyperref}

\def\confName{CVPR}
\def\confYear{2025}

\title{DepthSplat: Connecting Gaussian Splatting and Depth}

\author{Haofei Xu$^{1,2}$ \quad Songyou Peng$^{1}$ \quad  Fangjinhua Wang$^{1}$ \quad  Hermann Blum$^{1,4}$ \quad Daniel Barath$^{1}$ \\[4pt]
Andreas Geiger$^{2}$ \,  Marc Pollefeys$^{1,3}$ \\[4pt]
$^1$ETH Zurich \,  $^2$University of Tübingen, Tübingen AI Center \, $^3$Microsoft \, \, $^4$University of Bonn \\[4pt]
\href{https://haofeixu.github.io/depthsplat/}{haofeixu.github.io/depthsplat}
}

\begin{document}

\twocolumn[{
\renewcommand\twocolumn[1][]{#1}
\maketitle
\begin{center}
    \centering
    \vspace{-15pt}
    \includegraphics[width=\linewidth]{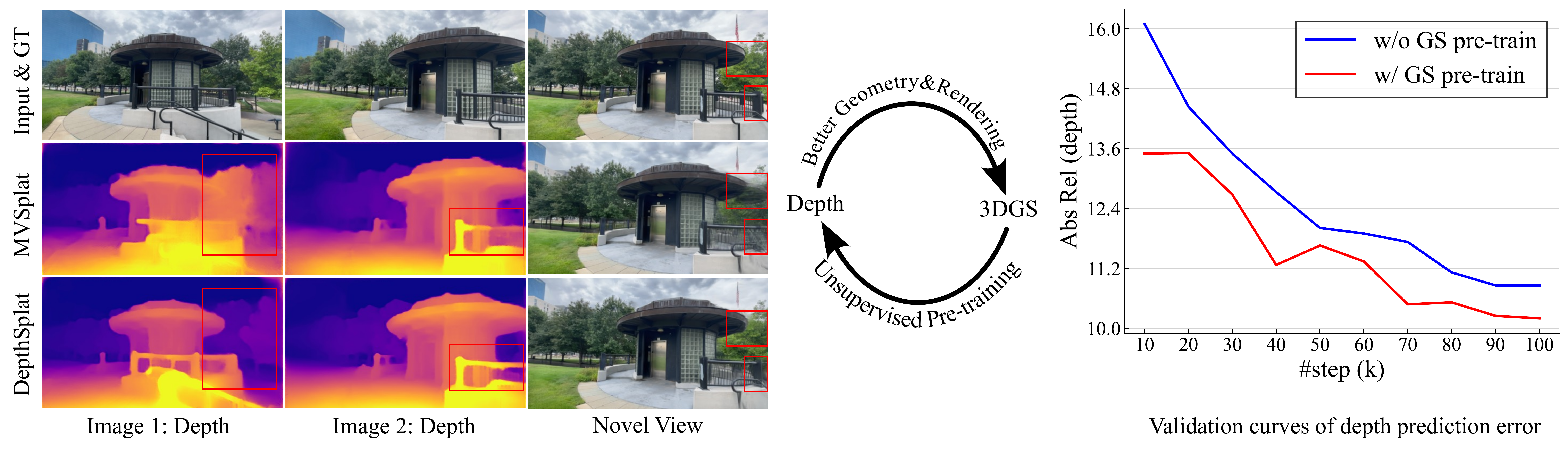}
    \vspace{-15pt}
    \captionof{figure}{\textbf{DepthSplat enables cross-task interactions between Gaussian splatting and depth}. \textit{Left:} Better depth leads to improved Gaussian splatting reconstruction. \textit{Right:} Unsupervised depth pre-training with Gaussian splatting leads to reduced depth prediction error. }
    \label{fig:teaser}
\end{center}
}]

\begin{abstract}
Gaussian splatting and single-view depth estimation are typically studied in isolation. In this paper, we present DepthSplat to connect Gaussian splatting and depth estimation and study their interactions. More specifically, we first contribute a robust multi-view depth model by leveraging pre-trained monocular depth features, leading to high-quality feed-forward 3D Gaussian splatting reconstructions. We also show that Gaussian splatting can serve as an unsupervised pre-training objective for learning powerful depth models from large-scale multi-view posed datasets. 
We validate the synergy between Gaussian splatting and depth estimation through extensive ablation and cross-task transfer experiments. Our DepthSplat achieves state-of-the-art performance on ScanNet, RealEstate10K and DL3DV datasets in terms of both depth estimation and novel view synthesis, demonstrating the mutual benefits of connecting both tasks. In addition, DepthSplat enables feed-forward reconstruction from 12 input views ($512 \times 960$ resolutions) in 0.6 seconds.
\vspace{-4mm}
\end{abstract}
    
\section{Introduction}
\label{sec:intro}

Novel view synthesis~\citep{buehler2001unstructured,zhou2018stereo} and depth prediction~\citep{schonberger2016pixelwise,eigen2014depth} are two fundamental tasks in computer vision, serving as the driving force behind numerous applications ranging from augmented reality to robotics and autonomous driving. There have been notable advancements in both areas recently.

For novel view synthesis, 3D Gaussian Splatting (3DGS)~\citep{Kerbl2023TOG} has emerged as a popular technique due to its impressive real-time performance while attaining high visual fidelity.
Recently, advances in feed-forward 3DGS models~\citep{charatan2023pixelsplat,Chen2024ARXIV,Szymanowicz2024CVPR} have been introduced to alleviate the need for tedious per-scene optimization, also enabling few-view 3D reconstruction.
The state-of-the-art sparse-view method MVSplat~\citep{Chen2024ARXIV} relies on feature matching-based multi-view depth estimation~\citep{xu2023unifying} to localize the 3D Gaussian positions, which makes it suffer from similar limitations (\emph{e.g.}, occlusions, texture-less regions, and reflective surfaces) as other multi-view depth methods~\citep{schonberger2016pixelwise,yao2018mvsnet,gu2020cascade,wang2021patchmatchnet,duzceker2021deepvideomvs}.

On the other hand, significant progress has
been made in monocular depth estimation, with recent models~\citep{Yang2024CVPR,Ke2024CVPR,Yin2023CVPR,Fu2024ARXIV,Ranftl2020TPAMI,Eftekhar2021ICCV}
achieving robust predictions on diverse in-the-wild data. However, these depths typically lack consistent scales across multiple views, constraining their performance in downstream tasks like 3D reconstruction~\citep{wang2023neural, yin2022towards}.

The integration of 3DGS with single-view depth estimation presents a compelling solution to overcome the individual limitations of each technique while at the same time enhancing their strengths. To this end, we introduce \emph{DepthSplat}, which exploits the complementary nature of sparse-view feed-forward 3DGS and robust monocular depth estimation to improve the performance for both tasks.

Specifically, we first contribute a robust multi-view depth model by integrating pre-trained monocular depth features~\citep{yang2024depth} to the multi-view feature matching branch, which not only maintains the consistency of multi-view depth models but also leads to more robust results in situations that are hard to match (\eg, occlusions, texture-less regions and reflective surfaces). The predicted multi-view depth maps are then unprojected to 3D as the Gaussian centers, and we use an additional lightweight network to predict other remaining Gaussian parameters. They are combined together to achieve novel view synthesis with the splatting operation~\citep{Kerbl2023TOG}.

\new{While previous methods~\citep{bae2022multi,li2023learning,cheng2024adaptive} also try to fuse monocular and multi-view depths, they usually rely on sophisticated architectures. 
In contrast, we identify the power of \textit{off-the-shelf} pre-trained monocular depth models and propose to augment multi-view cost volumes with monocular features, leading to a simpler model and stronger performance.}

Thanks to our improved multi-view depth model, the quality of novel view synthesis with Gaussian splatting is also significantly enhanced (see \cref{fig:teaser} left).
In addition, our Gaussian splatting module is fully differentiable, which requires only photometric supervision to optimize all model components. 
This provides a new, unsupervised way to pre-train depth prediction models on large-scale multi-view posed datasets without requiring ground truth geometry information. 
The pre-trained depth model can be further fine-tuned for specific depth tasks and achieves superior results over training from scratch (see \cref{fig:teaser} right, where unsupervised pre-training leads to improved performance).

We conduct extensive experiments on the large-scale TartanAir~\citep{wang2020tartanair}, ScanNet~\citep{dai2017scannet} and RealEstate10K~\citep{zhou2018stereo} datasets for depth estimation and Gaussian splatting tasks, as well as the recently introduced DL3DV~\citep{ling2024dl3dv} dataset, which features complex real-world scenes and thus is more challenging. Under various evaluation settings, our DepthSplat achieves state-of-the-art results. We also achieve promising feed-forward reconstruction results of large-scale or $360^{\circ}$ scenes from 12 input views ($512 \times 960$ resolutions) in 0.6 seconds. The strong performance on both tasks demonstrates the mutual benefits of connecting Gaussian splatting and depth.

\section{Related Work}
\label{sec:related_work}

\boldstartspace{Multi-View Depth Estimation}.
As a core component of classical multi-view stereo pipelines~\citep{schonberger2016pixelwise}, multi-view depth estimation exploits multi-view photometric consistency across multiple images to perform feature matching and predict the depth map of the reference image. In recent years, many learning-based methods~\citep{yao2018mvsnet,gu2020cascade,wang2021patchmatchnet,duzceker2021deepvideomvs,wang2021itermvs,ding2021transmvsnet,cao2022mvsformer} have been proposed to improve depth accuracy. 
Though these learning-based methods significantly improve the depth quality compared to traditional methods~\citep{galliani2015massively,schonberger2016pixelwise,xu2019acmm}, they cannot handle challenging situations where the multi-view photometric consistency assumption is not guaranteed,~\eg, occlusions, low-textured areas, and non-Lambertian surfaces.

\boldstartspace{Monocular Depth Estimation}. Recently, we have witnessed significant progress in depth estimation from a single image~\citep{Ranftl2020TPAMI,bhat2023zoedepth,Yin2023CVPR,Yang2024CVPR,Ke2024CVPR}, and existing methods can produce surprisingly accurate results on diverse in-the-wild data.
However, monocular depth methods inherently suffer from scale ambiguities, which makes it challenging to use them for downstream tasks like 3D reconstruction~\citep{yin2022towards}. In this paper, we leverage the powerful features from a pre-trained monocular depth model~\citep{yang2024depth} to augment feature-matching based multi-view depth estimation, which not only maintains the multi-view consistency but also leads to significantly improved robustness in challenging situations such as low-textured regions and reflective surfaces.

\begin{figure*}[!t]
    \centering
    \includegraphics[width=0.9\linewidth]{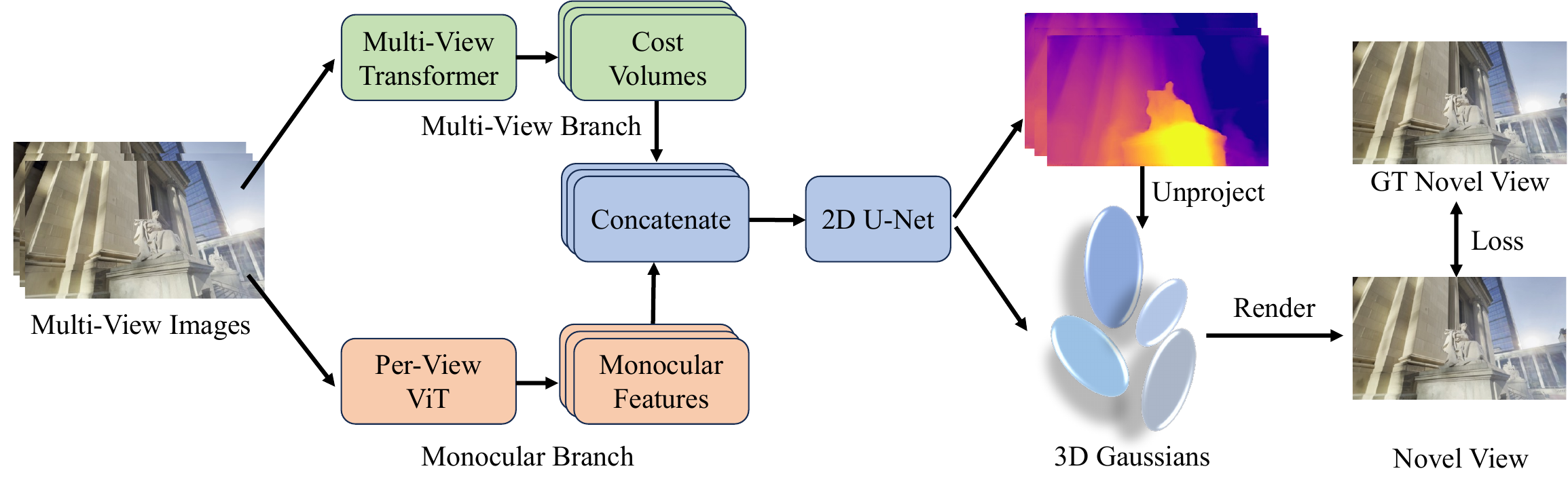}
    \caption{\textbf{DepthSplat} connects depth estimation and 3D Gaussian splatting with a shared architecture, which enables cross-task transfer. In particular, DepthSplat consists of a multi-view branch to model feature-matching information and a single-view branch to extract monocular features. The per-view cost volumes and monocular features are concatenated for depth regression with a 2D U-Net architecture. For the depth estimation task, we train the depth model with ground truth depth supervision. For the Gaussian splatting task, we first unproject all depth maps to 3D as the Gaussian centers, and in parallel, we use an additional head to predict the remaining Gaussian parameters. Novel views are rendered with the splatting operation. The full model for novel view synthesis is trained with the photometric rendering loss, which can also be used as an unsupervised pre-training stage for the depth model.
    }
    \label{fig:pipeline}
    \vspace{-4mm}
\end{figure*}

\boldstartspace{Feed-Forward Gaussian Splatting}. Several feed-forward 3D Gaussian splatting models~\citep{charatan2023pixelsplat,Szymanowicz2024CVPR,Chen2024ARXIV,wewer2024latentsplat,tang2024lgm,xu2024grm,zhang2024gs,chen2024mvsplat360,chen2024lara} have been proposed in literature thanks to its efficiency and ability to handle sparse views. In particular, pixelSplat~\citep{charatan2023pixelsplat} and Splatter Image~\citep{Szymanowicz2024CVPR} predict 3D Gaussians from image features, while MVSplat~\citep{Chen2024ARXIV} encodes the feature matching information with cost volumes and achieves better geometry. However, it inherently suffers from the limitation of feature matching in challenging situations like texture-less regions and reflective surfaces. In this paper, we propose to integrate monocular features from pre-trained monocular depth models~\cite{yang2024depth} for more robust depth prediction and 3D Gaussian reconstruction. Another line of work like LGM~\citep{tang2024lgm}, GRM~\citep{xu2024grm}, and GS-LRM~\citep{zhang2024gs} relies significantly on the training data and compute, discarding explicit feature matching cues and learning priors purely from data. This makes them expensive to train (\eg, GS-LRM~\citep{zhang2024gs} is trained with 64 A100 GPUs for 2 days), while our model can be trained in 2 days with 4 GPUs.
Moreover, our Gaussian splatting module enables pre-training depth model from large-scale multi-view posed datasets without ground truth depth supervision.

\boldstartspace{Depth and Gaussian Splatting}. \new{Flash3D~\cite{szymanowicz2024flash3d} explores a pre-trained monocular depth model~\cite{piccinelli2024unidepth} for single-image 3D Gaussian reconstruction. In contrast, we target large-scale scene reconstruction from multiple views. 
TranSplat~\cite{zhang2024transplat} leverages monocular depth to improve Gaussian reconstruction. However, a crucial difference is that TranSplat uses monocular features to refine the coarse depth predicted from a cost volume, which makes it inherently vulnerable to the error in the coarse stage and, thus, suffers from the error propagation issue. In contrast, we avoid this by combining cost volume and monocular features early, which performs significantly better than TranSplat (\cref{tab:sota_re10k}).} 
Another line of work~\citep{chung2024depth,turkulainen2024dn} applies an additional depth loss to the Gaussian optimization process. We note that these two approaches (feed-forward \emph{vs.} per-scene optimization) are orthogonal.

\section{DepthSplat}
\label{sec:method}

Given $N$ input images $\{{\bm I}^{i}\}_{i=1}^N, ({\bm I}^i \in \mathbb{R}^{H \times W \times 3}$, where $H$ and $W$ are the image sizes) with corresponding projection matrices $\{ \mathbf{P}_i \}_{i=1}^N$, ($\mathbf{P}_i \in \mathbb{R}^{3 \times 4}$, computed from the intrinsic and extrinsic matrices),
our goal is to predict dense per-pixel depth $\mathbf{D}_i \in \mathbb{R}^{H \times W}$ and per-pixel Gaussian parameters $\{ (\bm{\mu}_j, \alpha_j, \bm{\Sigma}_j, \bm{c}_j ) \}_{j=1}^{H \times W \times N}$ for each image, where $\bm{\mu}_j$, $\alpha_j$, $\bm{\Sigma}_j$ and $\bm{c}_j$ are the 3D Gaussian's position, opacity, covariance, and color. As shown in \cref{fig:pipeline}, key to our method is a multi-view depth model augmented with monocular depth features, where we obtain the position $\bm{\mu}_j$ of each Gaussian by unprojecting depth with camera parameters, and other Gaussian parameters are predicted by an additional head. 

More specifically, our depth model consists of two branches: one for modeling feature matching using cost volumes, and another for extracting monocular features from a pre-trained monocular depth network. The cost volumes and monocular features are concatenated together for subsequent depth regression with a 2D U-Net and a softmax layer. For the depth task, we train our depth model with ground truth depth supervision. Our full model for novel view synthesis is
trained with the rendering loss, which can also be used as an unsupervised pre-training stage for the depth model. We introduce the individual components below.

\subsection{Multi-View Feature Matching}
\label{sec:mv}

In this branch, we extract multi-view features with a multi-view Transformer architecture and then build multiple cost volumes that correspond to each input view.

\boldstartspace{Multi-View Feature Extraction}. For $N$ input images, we first use a lightweight weight-sharing ResNet~\citep{he2016deep} architecture to get $s\times$ downsampled features for each image independently. To handle different image resolutions, we make the downsampling factor $s$ flexible by controlling the number of stride-2 $3 \times 3$ convolutions. For example, the downsampling factor $s$ is 4 if two stride-2 convolutions are used and 8 if three are used.
To exchange information across different views, we use a multi-view Swin Transformer~\citep{liu2021swin,xu2022gmflow,xu2023unifying} which contains six stacked self- and cross-attention layers to obtain multi-view-aware features $\{ {\bm F_i} \}_{i=1}^N$, ${\bm F}^i \in \mathbb{R}^{\frac{H}{s} \times \frac{W}{s} \times C}$, where $C$ is the feature dimension. 
When more than two images ($N > 2$) are given as input, we perform cross-attention between each reference view and its top-$2$ nearest neighboring views, which are selected based on their camera distances to the reference view. This makes computation tractable with many input views.

\boldstartspace{Feature Matching}. We encode the feature matching information across different views with the plane-sweep stereo approach~\citep{collins1996space,xu2023unifying}. More specifically, for each view $i$, we first uniformly sample $D$ depth candidates $\{ d_{m} \}_{m=1}^D$ from the near and far depth ranges and then warp the feature ${\bm F}^j$ of view $j$ to view $i$ with the camera projection matrix and each depth candidate $d_m$. Then we obtain $D$ warped features $\{ {\bm F}^{j \to i}_{d_m} \}_{m=1}^D$ that correspond to feature ${\bm F}^i$. We then measure their feature correlations with the dot-product operation~\citep{xu2020aanet,Chen2024ARXIV}.
The cost volume ${\bm C}_i \in \mathbb{R}^{\frac{H}{s} \times \frac{W}{s} \times D}$ for image $i$ is obtained by stacking all correlations. Accordingly, we obtain cost volumes $\{ {\bm C}_i \}_{i=1}^N$ %
for all input images $\{ {\bm I}_i \}_{i=1}^N$.
For more than two input views, similar to the strategy in cross-view attention computation, we select the top-$2$ nearest views for each reference view and compute feature correlations with only the selected views. This enables our cost volume construction to achieve a good speed-accuracy trade-off and scale efficiently to a larger number of input views. The correlation values for the two selected views are combined with averaging.

\subsection{Monocular Depth Feature Extraction}

Despite the remarkable progress in multi-view feature matching-based depth estimation~\citep{yao2018mvsnet,wang2021itermvs,xu2023unifying} and Gaussian splatting~\citep{Chen2024ARXIV}, they inherently suffer from limitations in challenging situations like occlusions, texture-less regions, and reflective surfaces. 
Thus, we propose to integrate pre-trained monocular depth features into the cost volume to handle scenarios that are challenging or impossible to match. 

More specifically, we leverage the pre-trained monocular depth backbone from the recent Depth Anything V2~\citep{yang2024depth} model thanks to its impressive performance on diverse in-the-wild data. The monocular backbone is a ViT~\citep{dosovitskiy2020image,oquab2023dinov2} model, which has a patch size of 14 and outputs a feature map that is $1/14$ spatial resolution of the original image. 
We simply bilinearly interpolate the spatial resolution of the monocular features to the same resolution as the cost volume in \cref{sec:mv} and obtain the monocular feature ${\bm F}^i_{\mathrm{mono}} \in \mathbb{R}^{\frac{H}{s} \times \frac{W}{s} \times C_{\mathrm{mono}}}$ for input image ${\bm I}_i$, where $C_{\mathrm{mono}}$ is the dimension of the monocular feature. This process is performed for all input images in parallel and we obtain monocular features $\{ {\bm F}^i_{\mathrm{mono}} \in \mathbb{R}^{\frac{H}{s} \times \frac{W}{s} \times C_{\mathrm{mono}}} \}_{i=1}^N$, which are subsequently used for per-view depth map estimations.

\subsection{Feature Fusion and Depth Regression}
\label{sec:depth_regression}

To achieve robust and multi-view consistent depth predictions, we combine the monocular feature ${\bm F}^i_{\mathrm{mono}} \in \mathbb{R}^{\frac{H}{s} \times \frac{W}{s} \times C_{\mathrm{mono}}}$ and cost volume ${\bm C}_i \in \mathbb{R}^{\frac{H}{s} \times \frac{W}{s} \times D}$ via simple concatenation in the channel dimension. A subsequent 2D U-Net~\citep{rombach2022high,ronneberger2015u} is used to regress depth from the concatenated monocular features and cost volumes. This process is performed for all the input images in parallel and for each image, it outputs a tensor of shape $\frac{H}{s} \times \frac{W}{s} \times D$, where $D$ is the number of depth candidates. We normalize the $D$ dimension with the softmax operation and perform a weighted average of all depth candidates to obtain the depth output.

We also apply a hierarchical matching~\citep{gu2020cascade} architecture where an additional refinement step at $2\times$ higher feature resolution is employed to improve the performance further. More specifically, based on the coarse depth prediction, we perform a correspondence search on the $2\times$ higher feature maps within the neighbors of the $2\times$ upsampled coarse depth prediction. Since we already have a coarse depth prediction, we only need to search a smaller range at the higher resolution, and thus, we construct a smaller cost volume compared to the coarse stage. Such a $2$-scale hierarchical architecture not only leads to improved efficiency since most computation is spent on low resolution, but also leads to better results thanks to the use of higher-resolution features~\citep{gu2020cascade}. Similar feature fusion and depth regression procedure is used to get higher-resolution depth predictions, which are subsequently upsampled with a DPT head~\citep{ranftl2021vision} to the full resolution.

\subsection{Gaussian Parameter Prediction}

For the task of 3D Gaussian splatting, we directly unproject the per-pixel depth maps to 3D with the camera parameters as the Gaussian centers $\bm{\mu}_j$. To predict other remaining Gaussian parameters $\alpha_j$ (opacity), $\bm{\Sigma}_j$ (covariance) and $\bm{c}_j$ (color), we use an additional DPT head~\cite{ranftl2021vision}, where it takes the concatenated image, depth and feature information as input and outputs all the remaining Gaussian parameters. With all the predicted 3D Gaussians, we can render novel views with the Gaussian splatting operation~\citep{Kerbl2023TOG}.

\subsection{Training Loss}
\label{sec:loss}

We study the properties of our proposed model on two tasks: depth estimation and novel view synthesis with 3D Gaussian splatting~\citep{Kerbl2023TOG}. The loss functions are introduced below.

\boldstartspace{Depth estimation}. We train our depth model (without the Gaussian splatting head) with $\ell_1$ loss and gradient loss between the inverse depths of prediction and ground truth:
\begin{align}
    L_{\mathrm{depth}} &= \alpha \cdot |{\bm D}_{\mathrm{pred}} - {\bm D}_{\mathrm{gt}} |  + \beta \cdot |\partial_x {\bm D}_{\mathrm{pred}} - \partial_x {\bm D}_{\mathrm{gt}}| \notag \\
    &\quad + \beta \cdot | \partial_y {\bm D}_{\mathrm{pred}} - \partial_y {\bm D}_{\mathrm{gt}}|),
\label{loss:depth}
\end{align}
where $\partial_x$ and $\partial_y$ denotes the gradients on the $x$ and $y$ directions, respectively. Following UniMatch~\citep{xu2023unifying}, we use $\alpha=20$ and $\beta=20$.

\boldstartspace{View synthesis}. We train our full model with a combination of mean squared error (MSE) and LPIPS~\citep{zhang2018unreasonable} losses between rendered and ground truth image colors:
\begin{equation}
    L_{\mathrm{gs}} = \sum_{m=1}^M \left(\mathrm{MSE}(I^m_{\mathrm{render}}, I^m_{\mathrm{gt}}) + \lambda \cdot \mathrm{LPIPS}(I^m_{\mathrm{render}}, I^m_{\mathrm{gt}}) \right),
\label{loss:gs}
\end{equation}
where $M$ is the number of novel views to render in a single forward pass. 
The LPIPS loss weight $\lambda$ is set to 0.05~\cite{Chen2024ARXIV}.

\section{Experiments}
\label{sec:exp}

\boldstartspace{Implementation Details}. 
We implement our method in PyTorch~\citep{paszke2019pytorch} and optimize our model with the AdamW~\citep{loshchilov2017decoupled} optimizer and cosine learning rate schedule. We adopt the xFormers~\citep{lefaudeux2022xformers} library for our monocular ViT backbone implementation. We use a lower learning rate $2 \times 10^{-6}$ for the pre-trained Depth Anything V2~\cite{yang2024depth} backbone, and other remaining layers use a learning rate of $2 \times 10^{-4}$. The feature downsampling factor $s$ in our multi-view branch (\cref{sec:mv}) is chosen based on the image resolution. More specifically, for experiments on the $256 \times 256$ resolution RealEstate10K~\citep{zhou2018stereo} dataset, we choose $s=4$. For higher resolution datasets (\eg, TartanAir~\citep{wang2020tartanair}, ScanNet~\citep{dai2017scannet}, KITTI~\citep{geiger2013vision} and DL3DV~\citep{ling2024dl3dv}), we choose $s=8$. Our hierarchical matching models in \cref{sec:depth_regression} use 2-scale features, \ie, $1/8$ and $1/4$, or $1/4$ and $1/2$ resolutions.

\boldstartspace{Training Details}. For depth experiments on ScanNet~\cite{dai2017scannet}, we train our depth model on $4\times $ GH200 GPUs for 100K iterations with a total batch size of 32. For the ablation experiments on depth task, we mainly use synthetic datasets (TartanAir~\cite{wang2020tartanair} and VKITTI2~\cite{cabon2020virtual}) for training since they provide high-quality ground truth depths. For Gaussian splatting experiments, we consider both low-resolution and high-resolution images. In particular, for comparisons with previous methods~\cite{charatan2023pixelsplat,Chen2024ARXIV}, we mainly use $256\times 256$ resolution RealEstate10K~\citep{zhou2018stereo} and $256\times 448$ resolution DL3DV~\cite{ling2024dl3dv} datasets. We also report high-resolution ($512 \times 960$) results on RealEstate10K and DL3DV datasets for qualitative evaluations. For experiments on the $256\times 256$ RealEstate10K dataset, we train our model on $4 \times$ GH200 GPUs for 150K iterations with a total batch size of 32, which takes 1 day for the small model and 2 days for the large model. For experiments on the $256\times 448$ DL3DV~\citep{ling2024dl3dv} dataset, we fine-tune our RealEstate10K pre-trained model on $4 \times$ GH200 GPUs for 100K iterations with a total batch size of 4, where the number of input views is randomly sampled from 2 to 6. We evaluate the model's performance on different number of input views (2, 4, 6). For high-resolution results, we fine-tune our low-resolution pre-trained models on high-resolution images. More training details are provided in the appendix. 
Our code and training scripts are available at \href{https://github.com/cvg/depthsplat}{github.com/cvg/depthsplat} to ease reproducibility.

\begin{table}[!t]

\begin{center}
\footnotesize

\caption{\textbf{DepthSplat model variants}. We evaluate different monocular backbones and different multi-view models ($1$-scale and 2-scale features for hierarchical matching as described in \cref{sec:depth_regression}), where both larger monocular backbones and $2$-scale hierarchical models lead to consistently improved performance for both depth and view synthesis tasks. The inference time is measured on a single A100 GPU for two input views at $256 \times 256$ resolutions.
    }
    \vspace{-2mm}
    \setlength{\tabcolsep}{1.5pt} %
    \resizebox{\linewidth}{!}{
    \begin{tabular}{ccccccccccccccccccccccccc}
    \toprule
    \multirow{2}{*}[-2pt]{Mono} & \multirow{2}{*}[-2pt]{Multi-View} &  \multicolumn{2}{c}{Depth (TartanAir)} & \multicolumn{3}{c}{3DGS (RealEstate10K)} & \multirow{2}{*}[-2pt]{\begin{tabular}[x]{@{}c@{}}Param\\(M) \end{tabular}} & \multirow{2}{*}[-2pt]{\begin{tabular}[x]{@{}c@{}}Time\\(s) \end{tabular}}\\
    \cmidrule(lr){3-4} \cmidrule(lr){5-7} 
    \addlinespace[-12pt] \\
    & & Abs Rel $\downarrow$ & $\delta_1 \uparrow$ & PSNR $\uparrow$ & SSIM $\uparrow$ & LPIPS $\downarrow$ \\
    
    \midrule

    ViT-S & $1$-scale & 8.46 & 93.02 & 26.84 & 0.878 & 0.122 & 38 & 0.050 \\
    ViT-B & $1$-scale & 6.94 & 94.46 & 27.06 & 0.882 & 0.119 & 115 & 0.053 \\
    ViT-L & $1$-scale & \textbf{6.07} & \textbf{95.52} & \textbf{27.23} & \textbf{0.885} & \textbf{0.116} & 347 & 0.061 \\

    \midrule

    ViT-S & $2$-scale & 7.01 & 94.56 & 27.12 & 0.884 & 0.119 & 41 & 0.062 \\
    ViT-B & $2$-scale & 6.22 & 95.31 & 27.34 & 0.887 & 0.116 & 120 & 0.070 & \\
    ViT-L & $2$-scale & \textbf{5.57} & \textbf{96.07} & \textbf{27.47} & \textbf{0.889} & \textbf{0.114} & 354 & 0.079 \\

    \bottomrule
    \end{tabular}
    }

    \label{tab:depthsplat_models}
    \end{center}
    \vspace{-4mm}
\end{table}

\begin{table}[!t]

\begin{center}
\footnotesize

\caption{\textbf{Ablations}. We evaluate the contribution of the monocular feature branch and the cost volume branch (1st group of experiments), as well as different monocular features (2nd group of experiments). Our results indicate that the monocular feature and cost volume are complementary, with large performance drops when removing either one. The pre-trained Depth Anything V2 model weights achieve the best view synthesis results.
    }
    \vspace{-2mm}
    \setlength{\tabcolsep}{2.2pt} %
    \resizebox{\linewidth}{!}{
    \begin{tabular}{lcccccccccccccccccccccccc}
    \toprule
    \multirow{2}{*}[-2pt]{Components} &  \multicolumn{2}{c}{Depth (TartanAir)} & \multicolumn{3}{c}{3DGS (RealEstate10K)} \\ 
    \addlinespace[-12pt] \\
    \cmidrule(lr){2-3} \cmidrule(lr){4-6} 
    \addlinespace[-12pt] \\
    & Abs Rel $\downarrow$ & $\delta_1 \uparrow$ & PSNR $\uparrow$ & SSIM $\uparrow$ & LPIPS $\downarrow$ \\
    
    \midrule

    Full &  \textbf{8.46} & \textbf{93.02} & \textbf{26.84} & \textbf{0.878} & \textbf{0.122} \\
    w/o mono feature & 12.25 & 88.00 & 26.04 & 0.864 & 0.134 \\
    w/o cost volume & 11.34 & 90.02 & 23.24 & 0.766 & 0.184 \\

    \midrule

    ConvNeXt-T & 10.50 & 91.13 & 26.11 & 0.865 & 0.134 \\
    Midas & 9.53 & 91.61 & 26.46 & 0.872 & 0.128 \\
    DINO V2 & 8.93 & 92.49 & 26.76 & 0.877 & 0.123 \\
    Depth Anything V1 & \textbf{8.38} & \textbf{93.23} & 26.76 & 0.877 & 0.124 \\
    Depth Anything V2 & 8.46 & 93.02 & \textbf{26.84} & \textbf{0.878} & \textbf{0.122} \\

    \bottomrule
    \end{tabular}
    }

    \label{tab:ablation}
    \end{center}
    \vspace{-5mm}
\end{table}

\begin{figure*}[!t]
    \centering
    \includegraphics[width=\linewidth]{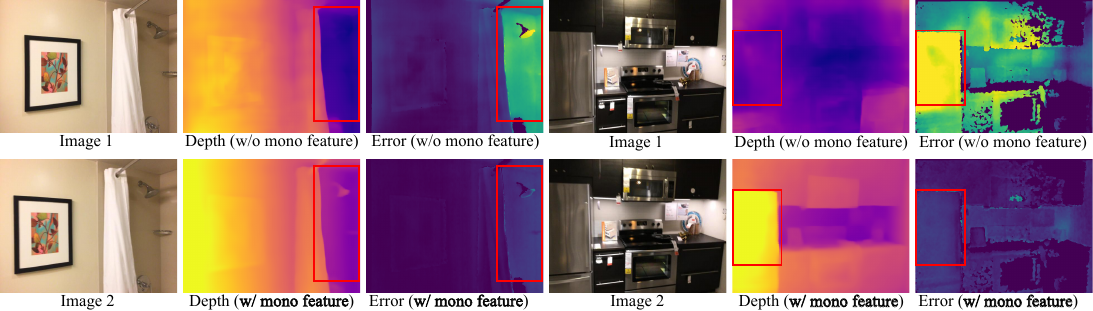}
    \vspace{-6mm}
    \caption{\textbf{Effect of monocular features for depth on ScanNet}. The monocular feature greatly improves challenging situations like texture-less regions (\emph{e.g.}, the wall in the first example) and reflective surfaces (\emph{e.g.}, the refrigerator in the second example).
    }
    \label{fig:depth_ablation}
\end{figure*}

\begin{figure*}[!t]
    \centering
    \includegraphics[width=\linewidth]{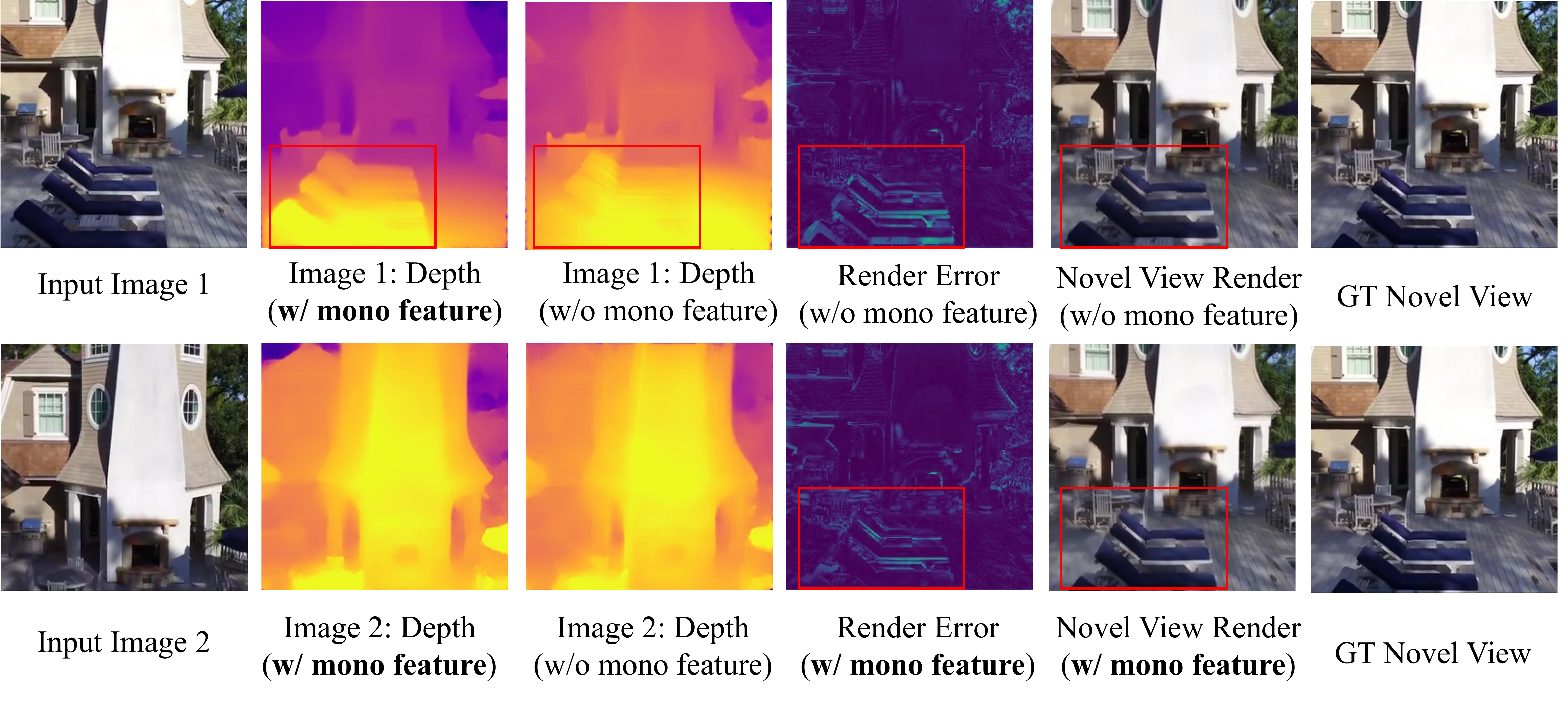}
    \vspace{-5mm}
    \caption{\textbf{Effect of monocular features for 3DGS on RealEstate10K}. Without monocular features, the model struggles at predicting reliable depth for pixels that are not able to find correspondences (\emph{e.g.}, the lounger chair highlighted with the read rectangle), which subsequently causes misalignment in the rendered image due to the incorrect geometry.
    }
    \label{fig:gs_ablation}
    \vspace{-3mm}
\end{figure*}

\subsection{Model Variants}
\label{sec:model_variants}

We first study several different model variants for both depth estimation and view synthesis tasks. In particular, we explore different monocular backbones~\citep{yang2024depth} (ViT-S, ViT-B, ViT-L) and different multi-view models ($1$-scale and $2$-scale). We conduct depth experiments on the large-scale TartanAir~\citep{wang2020tartanair} synthetic dataset, which features both indoor and outdoor scenes and has perfect ground truth depth. The Gaussian splatting experiments are conducted on the standard RealEstate10K~\citep{zhou2018stereo} dataset. Following community standards, we report the depth evaluation metrics~\citep{eigen2014depth} of Abs Rel (relative $\ell_1$ error) and $\delta_1$ (percentage of correctly estimated pixels within a threshold) and novel view synthesis metrics~\citep{Kerbl2023TOG} of PSNR, SSIM, and LPIPS. The results in \cref{tab:depthsplat_models} demonstrate that both larger monocular backbones and $2$-scale hierarchical models lead to consistently improved performance for both tasks. We name three different model sizes with ``Small": ``ViT-S monocular, 1-scale multi-view" , ``Base": ``ViT-B monocular, 2-scale multi-view", and ``Large": ``ViT-L monocular, 2-scale multi-view".

From \cref{tab:depthsplat_models}, we can also observe that better depth network architecture leads to improved view synthesis. In the appendix, we conduct additional experiments to study the effect of different initializations for the depth network to the view synthesis performance. Our results show that a better depth model initialization also contributes to improved rendering quality. In summary, both better depth network architecture and better depth model initialization lead to improved novel view synthesis results.

\subsection{Ablation and Analysis}

In this section, we study the properties of our key components on the TartanAir dataset (for depth) and RealEstate10K dataset (for view synthesis with Gaussian splatting). 

\boldstartspace{Monocular Features for Depth and Gaussian Splatting}. In \cref{tab:ablation}, we compare our full model (Full) with the model variant where the monocular depth feature branch (with a ViT-S model pre-trained by Depth Anything V2~\citep{yang2024depth}) is removed (w/o mono feature), leaving only the multi-view branch. We can observe a clear performance drop for both depth and view synthesis tasks. In \cref{fig:depth_ablation}, we visualize the depth predictions and error maps of both models on the ScanNet dataset. The pure multi-view feature matching-based approach struggles a lot at texture-less regions and reflective surfaces. At the same time, our full model achieves reliable results thanks to the powerful prior captured in monocular depth features. We also show the visual comparisons for the Gaussian splatting task in \cref{fig:gs_ablation} with two input views. For regions (\emph{e.g.}, the lounge chairs) that only appear in a single image, the pure multi-view method is unable to find correspondences. It thus produces unreliable depth predictions, leading to misalignment in the rendered novel views due to the incorrect geometry. 

We also experiment with removing the cost volume (w/o cost volume) in the multi-view branch and observed a significant performance drop. This indicates that obtaining scale- and multi-view consistent predictions with a pure monocular depth backbone is challenging, which constrains achieving high-quality results for 3D Gaussian reconstruction.

\begin{table}[!h]
    \begin{center}
\footnotesize
\caption{\textbf{Monocular fusion strategy}. Our simple concatenation performs surprisingly good compared to other alternatives. 
    }
    
    \setlength{\tabcolsep}{2.5pt} %
    \resizebox{0.7\linewidth}{!}{
    \begin{tabular}{lcccccccccccccccccccccccc}
    \toprule
    Method & Abs Rel $\downarrow$ & $\delta_1 \uparrow$ \\
        
    \toprule

    Cost volume from ViT & 11.26 & 90.84 \\
    Explicit scale align & 9.13 & 90.21 \\
    Attention fusion & \textbf{8.40} & 92.95 \\
    Concatenation (Ours) & 8.46 & \textbf{93.02} \\

    \bottomrule
    \end{tabular}
    }
    \end{center}
    
    \label{tab:mono_fusion}
    \vspace{-6mm}

\end{table}

\begin{table}[!t]
\begin{center}
\small
\caption{\new{\textbf{Unsupervised Depth Pre-Training with Gaussian Splatting}. 
We evaluate the same architecture but with different weights. The pre-training is performed on RealEstate10K, and fine-tuning is performed on TartanAir and VKITTI2. Unsupervised pre-training (2nd row) produces improved depth predictions compared to random initialization (1st). The performance can be further improved with additional fine-tuning (4th), where the benefit of pre-training is especially significant on challenging datasets like TartanAir and KITTI compared to training from random initialization (3rd).}
}
\vspace{-2mm}
\setlength{\tabcolsep}{2.2pt} %
\resizebox{\linewidth}{!}{
\begin{tabular}{ccccccccccccccccccccccc}
\toprule

\multirow{2}{*}[-2pt]{Pre-Train} & \multirow{2}{*}[-2pt]{Fine-Tune} &  \multicolumn{2}{c}{TartanAir} & \multicolumn{2}{c}{ScanNet} & \multicolumn{2}{c}{KITTI} \\ 
    \addlinespace[-12pt] \\
    \cmidrule(lr){3-4} \cmidrule(lr){5-6} \cmidrule(lr){7-8} 
    & & Abs Rel $\downarrow$ & $\delta_1 \uparrow$ & Abs Rel $\downarrow$ & $\delta_1 \uparrow$ & Abs Rel $\downarrow$ & $\delta_1 \uparrow$ \\
        
    \toprule

    \ding{55} & \ding{55} & 76.22 & 3.43 & 39.97 & 22.56 & 90.69 & 0.00 & \\

    \ding{51} & \ding{55} & 29.53 & 53.16 & 21.51 & 76.09 & 56.83 & 46.26 \\

    \ding{55} & \ding{51} & 10.86 & 90.55 & 6.70 & 96.14 & 11.56 & 87.27 \\

    \ding{51} & \ding{51} & \textbf{10.20} & \textbf{91.10} & \textbf{6.60} & \textbf{96.27} & \textbf{10.68} & \textbf{89.92} \\

\bottomrule
\end{tabular}
}
\label{tab:gs2depth}

\end{center}
\end{table}

\begin{table}[t]
\vspace{-2mm}

    \centering
    \caption{\textbf{Two-view Depth Estimation on ScanNet.} Our DepthSplat outperforms all prior methods by clear margins.}
    \vspace{-2mm}

    \setlength{\tabcolsep}{1pt} %
    \resizebox{\linewidth}{!}{
    \begin{tabular}{lccccccccccccccccccccccc}
    \toprule
    Method & Abs Rel $\downarrow$ & RMSE $\downarrow$ & RMSE$_\mathrm{log}$ $\downarrow$  \\
        
    \toprule
    
    DeMoN~\citep{ummenhofer2017demon} & 23.1 & 0.761 & 0.289 \\
    BA-Net~\citep{tang2018ba} & 16.1 & 0.346 & 0.214 \\
    DeepV2D~\citep{teed2019deepv2d} & 5.7 & 0.168 & 0.080 \\
    NeuralRecon~\citep{sun2021neuralrecon} & 4.7 & 0.164 & 0.093 \\
    DRO~\citep{gu2023dro} & 5.3 & 0.168 & 0.081 \\
    UniMatch~\citep{xu2023unifying} & 5.9 & 0.179 & 0.082 \\
    DepthSplat (w/o GS pre-train) & 4.5 & 0.125 & 0.061 \\
    DepthSplat (w/ GS pre-train) & \textbf{3.8} & \textbf{0.114} & \textbf{0.055} \\

    \bottomrule
\end{tabular}
}
\label{tab:scannet_depth}
\end{table}

\boldstartspace{Different Monocular Features}. In \cref{tab:ablation}, we also evaluate different monocular features, including the ConvNeXt-T~\citep{Liu2022CVPR} features used in AFNet~\citep{cheng2024adaptive}, and other popular monocular features including Midas~\citep{Ranftl2020TPAMI} and DINOv2~\citep{oquab2023dinov2}. The pre-trained Depth Anything V2~\cite{yang2024depth} monocular features achieve the best view synthesis results.

\boldstartspace{Monocular Fusion Strategy}. We compare with alternative strategies for fusing the monocular features for multi-view depth estimation in \cref{tab:mono_fusion}. In particular, we first compare with MVSFormer~\citep{cao2022mvsformer} that constructs the cost volume with monocular ViT features, which leads to a single-branch architecture. We can observe that our fusion strategy performs significantly better than the single-branch design, potentially because our two-branch design disentangles feature matching and monocular priors, which makes the learning task easier. We also compare with fusing the monocular depth maps (instead of monocular features in our model) by first aligning the monocular depth scale using the multi-view depth and then fusing the depth maps together with an additional network. Again, our simple concatenation performs better. We further compare our concatenation with the attention-based adaptive fusion and observe very similar results. Thus we opt to the simpler concatenation design.

\begin{table}
    \centering
    \caption{\textbf{Two-view 3DGS on RealEstate10K}.
Our DepthSplat achieves the best performance.}
\vspace{-2mm}
    \begin{tabular}{lccccccccccccccccccccccc}
    \toprule
    Method & PSNR $\uparrow$ & SSIM $\uparrow$ & LPIPS $\downarrow$  \\
        
    \toprule

    pixelNeRF~\citep{yu2021pixelnerf} & 20.43 & 0.589 & 0.550 \\
    GPNR~\citep{suhail2022generalizable} & 24.11 & 0.793 & 0.255 \\
    AttnRend~\citep{du2023learning} & 24.78 & 0.820 & 0.213 \\
    MuRF~\citep{xu2023murf} & 26.10 & 0.858 & 0.143 \\
    pixelSplat~\citep{charatan2023pixelsplat} & 25.89 & 0.858 & 0.142 \\
    MVSplat~\citep{Chen2024ARXIV} & 26.39 & 0.869 & 0.128 \\
    TranSplat~\cite{zhang2024transplat} & 26.69 & 0.875 & 0.125 \\
    DepthSplat & \textbf{27.47} & \textbf{0.889} & \textbf{0.114} \\
    
    \bottomrule
    \end{tabular}

    \label{tab:sota_re10k}
\end{table}

\begin{figure}[t]
    \centering
    \includegraphics[width=\linewidth]{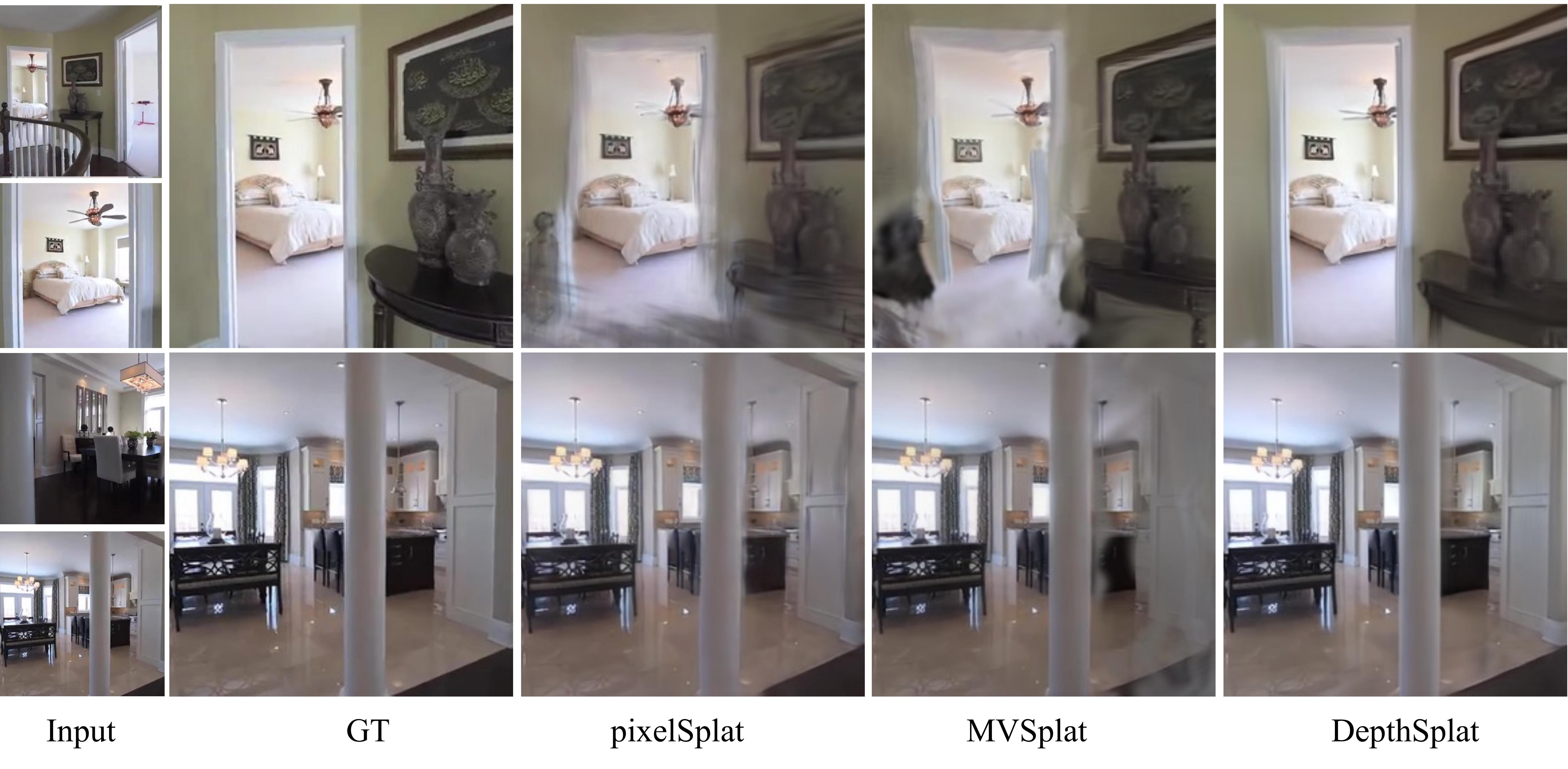}
    \vspace{-6mm}
    \caption{\textbf{Novel view synthesis on RealEstate10K}. Our DepthSplat performs significantly better than pixelSplat~\citep{charatan2023pixelsplat} and MVSplat~\citep{Chen2024ARXIV} in challenging regions.
    }
    \label{fig:gs_re10k}
\end{figure}

\subsection{Unsupervised Depth Pre-Training with Gaussian Splatting}

By connecting Gaussian splatting and depth, our DepthSplat provides a way to pre-train the depth model in a fully unsupervised manner. In particular, we first train our full model on the large-scale multi-view posed RealEstate10K dataset (contains $\sim$ 67K Youtube videos) with only the Gaussian splatting rendering loss (\cref{loss:gs}), without any direct supervision on the depth predictions. After pre-training, we take the pre-trained depth model and further fine-tune it to the depth task on the mixed TartanAir~\cite{wang2020tartanair} and VKITTI2~\citep{cabon2020virtual} datasets with ground truth depth supervision. In \cref{tab:gs2depth}, we evaluate the performance on both in-domain TartanAir test set and the zero-shot generalization on unseen ScanNet and KITTI datasets. We can observe that the benefit of pre-training is especially significant on the challenging datasets like TartanAir and KITTI. The results align well with the observation in~\cite{erhan2010does} which suggests that the pre-training acts as regularization and guides the learning towards better minima and enables better generalization to out-of-distribution datasets. The visual comparison results are provided in the appendix. Given the increasing popularity of view synthesis~\citep{zheng2023free3d,weng2023consistent123} and multi-view generative models~\citep{shi2023mvdream,blattmann2023stable}, new multi-view datasets~\citep{ling2024dl3dv} and models~\citep{voleti2024sv3d} are gradually introduced, our approach provides a way to pre-train depth models on large-scale multi-view posed image datasets.

\subsection{Benchmark Comparisons}

\boldstartspace{Comparisons on ScanNet and RealEstate10K}. \cref{tab:scannet_depth} and \cref{tab:sota_re10k} compare the depth and novel view synthesis results on the standard ScanNet and RealEstate10K benchmarks, respectively. We can see clearly that our DepthSplat achieves state-of-the-art performance on both datasets for both tasks. The visual comparison with previous methods is shown in \cref{fig:gs_re10k}, where our method significantly improves the performance on challenging scenarios like texture-less regions and occlusions. For the depth results on ScanNet, we report our base models (with ViT-B and two-scale feature matching) trained with and without Gaussian splatting pre-training on the mixed RealEstate10K and DL3DV datasets. Our model significantly outperforms previous methods and can benefit from additional Gaussian splatting pre-training.

\begin{table}[t]

\begin{center}
\small
\caption{\textbf{Comparisons on DL3DV}. 
Our DepthSplat not only consistently outperforms MVSplat on different number of input views, but also scales more efficiently to more input views.
}
\resizebox{\linewidth}{!}{
\begin{tabular}{lccccccccccccccccccccccc}
\toprule
Method & \#Views & PSNR $\uparrow$ & SSIM $\uparrow$ & LPIPS $\downarrow$ & Time (s) \\
    
\toprule

MVSplat~\cite{Chen2024ARXIV} & \multirow{2}{*}[-2pt]{2} & 17.54 & 0.529 & 0.402 & \textbf{0.072} \\
DepthSplat & & \textbf{19.31} & \textbf{0.615} & \textbf{0.310} & 0.083 \\

\midrule

MVSplat~\cite{Chen2024ARXIV} & \multirow{2}{*}[-2pt]{4} & 21.63 & 0.721 & 0.233 & 0.146 \\
DepthSplat & & \textbf{23.12} & \textbf{0.780} & \textbf{0.178} & \textbf{0.107} \\

\midrule

MVSplat~\cite{Chen2024ARXIV} & \multirow{2}{*}[-2pt]{6} & 22.93 & 0.775 & 0.193 & 0.263 \\
DepthSplat & & \textbf{24.19} & \textbf{0.823} & \textbf{0.147} & \textbf{0.132} \\

\bottomrule
\end{tabular}
\label{tab:dl3dv_compare}

}
\end{center}
\end{table}

\begin{figure}[!t]
    \centering
    \includegraphics[width=\linewidth]{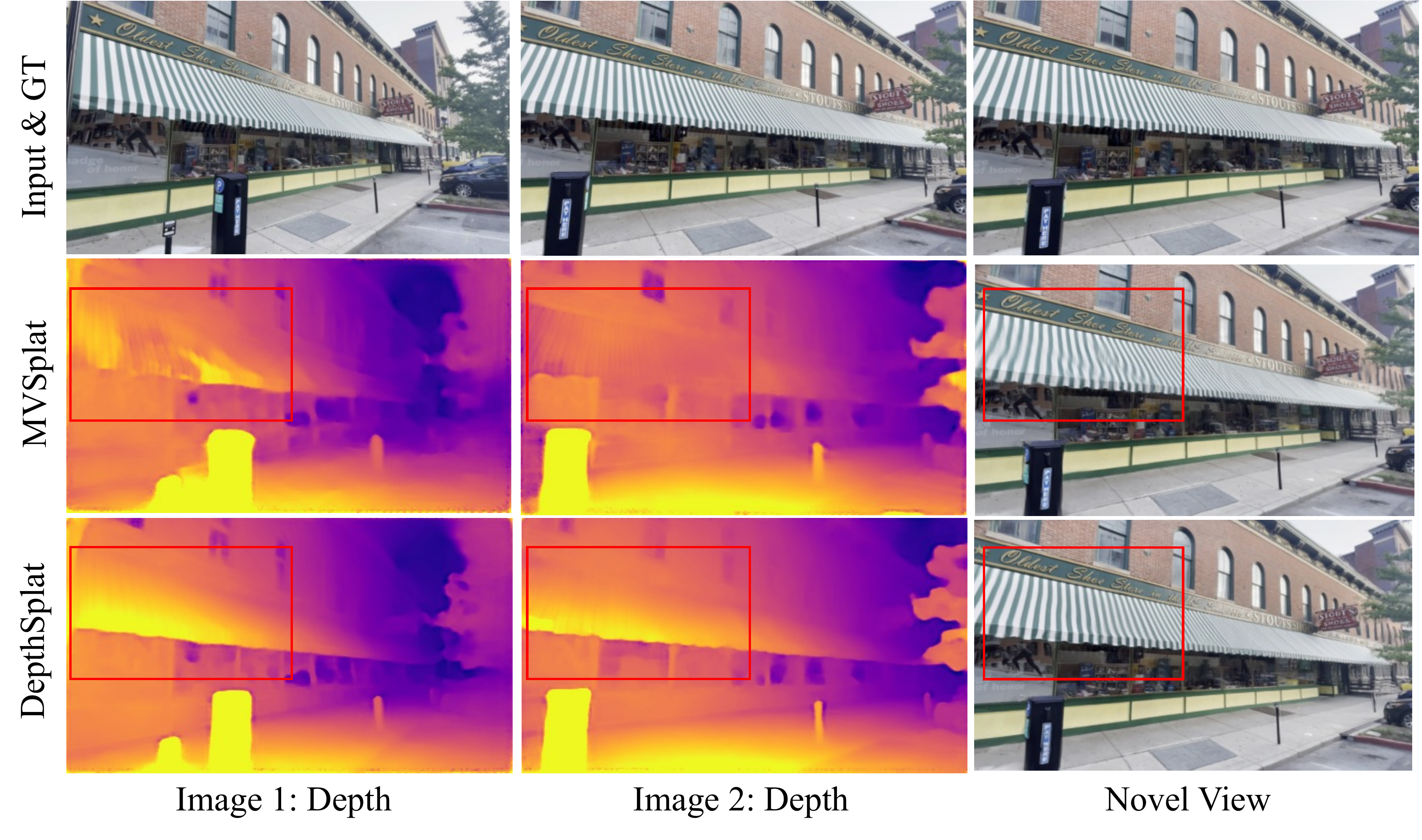}
    \caption{\textbf{Visual comparisons on DL3DV}. Our DepthSplat performs significantly better than MVSplat~\citep{Chen2024ARXIV} on regions that hard to match (\eg, repeated patterns).
    }
    \label{fig:gs_dl3dv}
\end{figure}

\boldstartspace{Comparisons on DL3DV}. To evaluate the performance on complex real-world scenes, we conduct comparisons with the representative model MVSplat~\citep{Chen2024ARXIV} on the recently introduced DL3DV dataset~\citep{ling2024dl3dv}. We also compare the results of different numbers of input views (2, 4 and 6) on this dataset. We fine-tune MVSplat and our RealEstate10K pre-trained models on DL3DV training scenes and report the results on the benchmark test set in \cref{tab:dl3dv_compare}. Our DepthSplat consistently outperforms MVSplat in all metrics, and our improvements are more significant with fewer input views, which indicates that our model is more robust to sparse views. Visual comparisons with MVSplat on the DL3DV dataset are shown in \cref{fig:gs_dl3dv}, where MVSplat's depth quality lags behind our DepthSplat due to matching failure, which leads to blurry and distorted view synthesis results. We show more visual comparison results in the appendix. It is also worth noting that our method scales more efficiently to more input views thanks to our lightweight local feature matching approach (\cref{sec:mv}), which is unlike the expensive global pair-wise matching used in MVSplat.

\begin{table}[t]

\begin{center}
\footnotesize

\caption{\new{\textbf{Cross-dataset generalization}. When generalizing to DL3DV and ACID with RealEstate10K pre-trained models, our DepthSplat is consistently better than MVSplat, especially on the more challenging DL3DV dataset.}
    }
    \vspace{-2mm}
    \setlength{\tabcolsep}{2.2pt} %
    \resizebox{\linewidth}{!}{
    \begin{tabular}{lcccccccccccccccccccccccc}
    \toprule
    \multirow{2}{*}[-2pt]{Method} &  \multicolumn{3}{c}{DL3DV} & \multicolumn{3}{c}{ACID} \\ 
    \addlinespace[-12pt] \\
    \cmidrule(lr){2-4} \cmidrule(lr){5-7} 
    \addlinespace[-12pt] \\
    & PSNR $\uparrow$ & SSIM $\uparrow$ & LPIPS $\downarrow$ & PSNR $\uparrow$ & SSIM $\uparrow$ & LPIPS $\downarrow$ \\
    
    \midrule

    MVSplat~\cite{Chen2024ARXIV} &  24.14 & 0.808 & 0.149 & 28.15 & 0.841 & 0.147 \\
    DepthSplat & \textbf{27.66} & \textbf{0.895} & \textbf{0.096} & \textbf{28.37} & \textbf{0.847} & \textbf{0.141} \\

    \bottomrule
    \end{tabular}
    }

    \label{tab:generalization}
    \end{center}
    \vspace{-6mm}
\end{table}

\boldstartspace{Cross-Dataset Generalization}. We evaluate the cross-dataset generalization capability by directly testing on DL3DV~\cite{ling2024dl3dv} and ACID~\cite{liu2021infinite} datasets with RealEstate10K~\cite{zhou2018stereo} pre-trained models. The results are shown in \cref{tab:generalization}, where our DepthSplat consistently outperforms MVSplat on both datasets, especially on the more challenging DL3DV dataset which features complex scene structures. The visual comparison results are shown in the appendix.

\boldstartspace{High-Resolution Results.} In addition to the comparison results on the $256 \times 256$ resolution RealEstate10K and $256\times 448$ resolution DL3DV dataets, we show more high-resolution ($512 \times 960$) qualitative results in the appendix. We also invite the readers to our project page \href{https://haofeixu.github.io/depthsplat/}{haofeixu.github.io/depthsplat} for high-resolution ($512 \times 960$) video results on different number of input views (6 and 12), where our DepthSplat is able to reconstruct larger-scale or $360^{\circ}$ scenes from more input views.

\section{Conclusion}
\label{sec:conclusion}

In this paper, we introduce DepthSplat, a new approach to connecting Gaussian splatting and depth to achieve state-of-the-art results on ScanNet, RealEstate10K and DL3DV datasets for both depth and view synthesis tasks. 
We show that our model enables unsupervised depth pre-training with Gaussian splatting rendering loss, providing a way to leverage large-scale posed multi-view image datasets for training more robust and generalizable multi-view depth models. 

\boldstartspace{Limitations}. Our current model requires camera poses as input along with the multi-view images, which might be challenging to obtain when the input views are extremely sparse. An interesting future direction is to remove this requiement by exploring pose-free models~\cite{ye2024no,zhang2025flare}. In addition, our model predicts pixel-aligned Gaussians, where the number of Gaussians will be significantly large when processing a large number of input views~\cite{wang2024freesplat,zhang2024gaussian}. Further improving the geometry representation and the scalability to many input views would be exciting future work.

\section*{Acknowledgement}
\label{sec:acknowledgement}

We thank Yuedong Chen for his generous help with the DL3DV dataset. Andreas Geiger was supported by the ERC Starting Grant LEGO-3D (850533) and the DFG EXC number 2064/1 - project number 390727645. This work was supported as part of the Swiss AI Initiative by a grant from the Swiss National Supercomputing Centre (CSCS) under project ID a03 on Alps. H. Blum is also with the Lamarr Institute and D. Barath is also with HUN-REN SZTAKI.

{
    \small
    \bibliographystyle{ieeenat_fullname}
    \bibliography{main}

\begin{thebibliography}{82}
\providecommand{\natexlab}[1]{#1}
\providecommand{\url}[1]{\texttt{#1}}
\expandafter\ifx\csname urlstyle\endcsname\relax
  \providecommand{\doi}[1]{doi: #1}\else
  \providecommand{\doi}{doi: \begingroup \urlstyle{rm}\Url}\fi

\bibitem[Bae et~al.(2022)Bae, Budvytis, and Cipolla]{bae2022multi}
Gwangbin Bae, Ignas Budvytis, and Roberto Cipolla.
\newblock Multi-view depth estimation by fusing single-view depth probability with multi-view geometry.
\newblock In \emph{CVPR}, 2022.

\bibitem[Bhat et~al.(2023)Bhat, Alhashim, and Wonka]{bhat2023zoedepth}
Shariq~Farooq Bhat, Ibraheem Alhashim, and Peter Wonka.
\newblock Zoedepth: Zero-shot transfer by combining relative and metric depth.
\newblock In \emph{CVPR}, 2023.

\bibitem[Blattmann et~al.(2023)Blattmann, Dockhorn, Kulal, Mendelevitch, Kilian, Lorenz, Levi, English, Voleti, Letts, et~al.]{blattmann2023stable}
Andreas Blattmann, Tim Dockhorn, Sumith Kulal, Daniel Mendelevitch, Maciej Kilian, Dominik Lorenz, Yam Levi, Zion English, Vikram Voleti, Adam Letts, et~al.
\newblock Stable video diffusion: Scaling latent video diffusion models to large datasets.
\newblock 2023.

\bibitem[Buehler et~al.(2001)Buehler, Bosse, McMillan, Gortler, and Cohen]{buehler2001unstructured}
Chris Buehler, Michael Bosse, Leonard McMillan, Steven Gortler, and Michael Cohen.
\newblock Unstructured lumigraph rendering.
\newblock In \emph{ACM TOG}, 2001.

\bibitem[Cabon et~al.(2020)Cabon, Murray, and Humenberger]{cabon2020virtual}
Yohann Cabon, Naila Murray, and Martin Humenberger.
\newblock Virtual kitti 2.
\newblock \emph{arXiv preprint arXiv:2001.10773}, 2020.

\bibitem[Cao et~al.(2022)Cao, Ren, and Fu]{cao2022mvsformer}
Chenjie Cao, Xinlin Ren, and Yanwei Fu.
\newblock Mvsformer: Learning robust image representations via transformers and temperature-based depth for multi-view stereo.
\newblock 2022.

\bibitem[Charatan et~al.(2024)Charatan, Li, Tagliasacchi, and Sitzmann]{charatan2023pixelsplat}
David Charatan, Sizhe Li, Andrea Tagliasacchi, and Vincent Sitzmann.
\newblock pixelsplat: 3d gaussian splats from image pairs for scalable generalizable 3d reconstruction.
\newblock In \emph{CVPR}, 2024.

\bibitem[Chen et~al.(2024{\natexlab{a}})Chen, Xu, Esposito, Tang, and Geiger]{chen2024lara}
Anpei Chen, Haofei Xu, Stefano Esposito, Siyu Tang, and Andreas Geiger.
\newblock Lara: Efficient large-baseline radiance fields.
\newblock In \emph{ECCV}, 2024{\natexlab{a}}.

\bibitem[Chen et~al.(2024{\natexlab{b}})Chen, Xu, Zheng, Zhuang, Pollefeys, Geiger, Cham, and Cai]{Chen2024ARXIV}
Yuedong Chen, Haofei Xu, Chuanxia Zheng, Bohan Zhuang, Marc Pollefeys, Andreas Geiger, Tat-Jen Cham, and Jianfei Cai.
\newblock Mvsplat: Efficient 3d gaussian splatting from sparse multi-view images.
\newblock In \emph{ECCV}, 2024{\natexlab{b}}.

\bibitem[Chen et~al.(2024{\natexlab{c}})Chen, Zheng, Xu, Zhuang, Vedaldi, Cham, and Cai]{chen2024mvsplat360}
Yuedong Chen, Chuanxia Zheng, Haofei Xu, Bohan Zhuang, Andrea Vedaldi, Tat-Jen Cham, and Jianfei Cai.
\newblock Mvsplat360: Feed-forward 360 scene synthesis from sparse views.
\newblock In \emph{NeurIPS}, 2024{\natexlab{c}}.

\bibitem[Cheng et~al.(2024)Cheng, Yin, Wang, Chen, Wang, and Yang]{cheng2024adaptive}
JunDa Cheng, Wei Yin, Kaixuan Wang, Xiaozhi Chen, Shijie Wang, and Xin Yang.
\newblock Adaptive fusion of single-view and multi-view depth for autonomous driving.
\newblock In \emph{CVPR}, 2024.

\bibitem[Chung et~al.(2024)Chung, Oh, and Lee]{chung2024depth}
Jaeyoung Chung, Jeongtaek Oh, and Kyoung~Mu Lee.
\newblock Depth-regularized optimization for 3d gaussian splatting in few-shot images.
\newblock In \emph{CVPR}, 2024.

\bibitem[Collins(1996)]{collins1996space}
Robert~T Collins.
\newblock A space-sweep approach to true multi-image matching.
\newblock In \emph{CVPR}, 1996.

\bibitem[Dai et~al.(2017)Dai, Chang, Savva, Halber, Funkhouser, and Nie{\ss}ner]{dai2017scannet}
Angela Dai, Angel~X Chang, Manolis Savva, Maciej Halber, Thomas Funkhouser, and Matthias Nie{\ss}ner.
\newblock Scannet: Richly-annotated 3d reconstructions of indoor scenes.
\newblock In \emph{CVPR}, 2017.

\bibitem[Ding et~al.(2022)Ding, Yuan, Zhu, Zhang, Liu, Wang, and Liu]{ding2021transmvsnet}
Yikang Ding, Wentao Yuan, Qingtian Zhu, Haotian Zhang, Xiangyue Liu, Yuanjiang Wang, and Xiao Liu.
\newblock Transmvsnet: Global context-aware multi-view stereo network with transformers.
\newblock \emph{CVPR}, 2022.

\bibitem[Dosovitskiy et~al.(2021)Dosovitskiy, Beyer, Kolesnikov, Weissenborn, Zhai, Unterthiner, Dehghani, Minderer, Heigold, Gelly, et~al.]{dosovitskiy2020image}
Alexey Dosovitskiy, Lucas Beyer, Alexander Kolesnikov, Dirk Weissenborn, Xiaohua Zhai, Thomas Unterthiner, Mostafa Dehghani, Matthias Minderer, Georg Heigold, Sylvain Gelly, et~al.
\newblock An image is worth 16x16 words: Transformers for image recognition at scale.
\newblock 2021.

\bibitem[Du et~al.(2023)Du, Smith, Tewari, and Sitzmann]{du2023learning}
Yilun Du, Cameron Smith, Ayush Tewari, and Vincent Sitzmann.
\newblock Learning to render novel views from wide-baseline stereo pairs.
\newblock In \emph{CVPR}, 2023.

\bibitem[Duzceker et~al.(2021)Duzceker, Galliani, Vogel, Speciale, Dusmanu, and Pollefeys]{duzceker2021deepvideomvs}
Arda Duzceker, Silvano Galliani, Christoph Vogel, Pablo Speciale, Mihai Dusmanu, and Marc Pollefeys.
\newblock Deepvideomvs: Multi-view stereo on video with recurrent spatio-temporal fusion.
\newblock In \emph{CVPR}, 2021.

\bibitem[Eftekhar et~al.(2021)Eftekhar, Sax, Malik, and Zamir]{Eftekhar2021ICCV}
Ainaz Eftekhar, Alexander Sax, Jitendra Malik, and Amir Zamir.
\newblock Omnidata: A scalable pipeline for making multi-task mid-level vision datasets from 3d scans.
\newblock In \emph{ICCV}, 2021.

\bibitem[Eigen et~al.(2014)Eigen, Puhrsch, and Fergus]{eigen2014depth}
David Eigen, Christian Puhrsch, and Rob Fergus.
\newblock Depth map prediction from a single image using a multi-scale deep network.
\newblock 2014.

\bibitem[Erhan et~al.(2010)Erhan, Courville, Bengio, and Vincent]{erhan2010does}
Dumitru Erhan, Aaron Courville, Yoshua Bengio, and Pascal Vincent.
\newblock Why does unsupervised pre-training help deep learning?
\newblock In \emph{JMLR}, 2010.

\bibitem[Fu et~al.(2024)Fu, Yin, Hu, Wang, Ma, Tan, Shen, Lin, and Long]{Fu2024ARXIV}
Xiao Fu, Wei Yin, Mu Hu, Kaixuan Wang, Yuexin Ma, Ping Tan, Shaojie Shen, Dahua Lin, and Xiaoxiao Long.
\newblock Geowizard: Unleashing the diffusion priors for 3d geometry estimation from a single image.
\newblock 2024.

\bibitem[Galliani et~al.(2015)Galliani, Lasinger, and Schindler]{galliani2015massively}
Silvano Galliani, Katrin Lasinger, and Konrad Schindler.
\newblock Massively parallel multiview stereopsis by surface normal diffusion.
\newblock In \emph{ICCV}, 2015.

\bibitem[Geiger et~al.(2013)Geiger, Lenz, Stiller, and Urtasun]{geiger2013vision}
Andreas Geiger, Philip Lenz, Christoph Stiller, and Raquel Urtasun.
\newblock Vision meets robotics: The kitti dataset.
\newblock \emph{The International Journal of Robotics Research}, 32\penalty0 (11):\penalty0 1231--1237, 2013.

\bibitem[Gu et~al.(2020)Gu, Fan, Zhu, Dai, Tan, and Tan]{gu2020cascade}
Xiaodong Gu, Zhiwen Fan, Siyu Zhu, Zuozhuo Dai, Feitong Tan, and Ping Tan.
\newblock Cascade cost volume for high-resolution multi-view stereo and stereo matching.
\newblock In \emph{CVPR}, 2020.

\bibitem[Gu et~al.(2023)Gu, Yuan, Dai, Zhu, Tang, Dong, and Tan]{gu2023dro}
Xiaodong Gu, Weihao Yuan, Zuozhuo Dai, Siyu Zhu, Chengzhou Tang, Zilong Dong, and Ping Tan.
\newblock Dro: Deep recurrent optimizer for video to depth.
\newblock 2023.

\bibitem[He et~al.(2016)He, Zhang, Ren, and Sun]{he2016deep}
Kaiming He, Xiangyu Zhang, Shaoqing Ren, and Jian Sun.
\newblock Deep residual learning for image recognition.
\newblock In \emph{CVPR}, 2016.

\bibitem[Ke et~al.(2024)Ke, Obukhov, Huang, Metzger, Daudt, and Schindler]{Ke2024CVPR}
Bingxin Ke, Anton Obukhov, Shengyu Huang, Nando Metzger, Rodrigo~Caye Daudt, and Konrad Schindler.
\newblock Repurposing diffusion-based image generators for monocular depth estimation.
\newblock In \emph{CVPR}, 2024.

\bibitem[Kerbl et~al.(2023)Kerbl, Kopanas, Leimk{\"u}hler, and Drettakis]{Kerbl2023TOG}
Bernhard Kerbl, Georgios Kopanas, Thomas Leimk{\"u}hler, and George Drettakis.
\newblock 3d gaussian splatting for real-time radiance field rendering.
\newblock \emph{ACM TOG}, 2023.

\bibitem[Lefaudeux et~al.(2022)Lefaudeux, Massa, Liskovich, Xiong, Caggiano, Naren, Xu, Hu, Tintore, Zhang, et~al.]{lefaudeux2022xformers}
Benjamin Lefaudeux, Francisco Massa, Diana Liskovich, Wenhan Xiong, Vittorio Caggiano, Sean Naren, Min Xu, Jieru Hu, Marta Tintore, Susan Zhang, et~al.
\newblock xformers: A modular and hackable transformer modelling library, 2022.

\bibitem[Li et~al.(2023)Li, Gong, Yin, Chen, Zhu, Wang, Chen, Sun, and Zhang]{li2023learning}
Rui Li, Dong Gong, Wei Yin, Hao Chen, Yu Zhu, Kaixuan Wang, Xiaozhi Chen, Jinqiu Sun, and Yanning Zhang.
\newblock Learning to fuse monocular and multi-view cues for multi-frame depth estimation in dynamic scenes.
\newblock In \emph{CVPR}, 2023.

\bibitem[Ling et~al.(2024)Ling, Sheng, Tu, Zhao, Xin, Wan, Yu, Guo, Yu, Lu, et~al.]{ling2024dl3dv}
Lu Ling, Yichen Sheng, Zhi Tu, Wentian Zhao, Cheng Xin, Kun Wan, Lantao Yu, Qianyu Guo, Zixun Yu, Yawen Lu, et~al.
\newblock Dl3dv-10k: A large-scale scene dataset for deep learning-based 3d vision.
\newblock In \emph{CVPR}, 2024.

\bibitem[Liu et~al.(2021{\natexlab{a}})Liu, Tucker, Jampani, Makadia, Snavely, and Kanazawa]{liu2021infinite}
Andrew Liu, Richard Tucker, Varun Jampani, Ameesh Makadia, Noah Snavely, and Angjoo Kanazawa.
\newblock Infinite nature: Perpetual view generation of natural scenes from a single image.
\newblock In \emph{ICCV}, 2021{\natexlab{a}}.

\bibitem[Liu et~al.(2021{\natexlab{b}})Liu, Lin, Cao, Hu, Wei, Zhang, Lin, and Guo]{liu2021swin}
Ze Liu, Yutong Lin, Yue Cao, Han Hu, Yixuan Wei, Zheng Zhang, Stephen Lin, and Baining Guo.
\newblock Swin transformer: Hierarchical vision transformer using shifted windows.
\newblock In \emph{ICCV}, 2021{\natexlab{b}}.

\bibitem[Liu et~al.(2022)Liu, Mao, Wu, Feichtenhofer, Darrell, and Xie]{Liu2022CVPR}
Zhuang Liu, Hanzi Mao, Chao-Yuan Wu, Christoph Feichtenhofer, Trevor Darrell, and Saining Xie.
\newblock A convnet for the 2020s.
\newblock In \emph{CVPR}, 2022.

\bibitem[Loshchilov and Hutter(2019)]{loshchilov2017decoupled}
Ilya Loshchilov and Frank Hutter.
\newblock Decoupled weight decay regularization.
\newblock 2019.

\bibitem[Oquab et~al.(2023)Oquab, Darcet, Moutakanni, Vo, Szafraniec, Khalidov, Fernandez, Haziza, Massa, El-Nouby, et~al.]{oquab2023dinov2}
Maxime Oquab, Timoth{\'e}e Darcet, Th{\'e}o Moutakanni, Huy Vo, Marc Szafraniec, Vasil Khalidov, Pierre Fernandez, Daniel Haziza, Francisco Massa, Alaaeldin El-Nouby, et~al.
\newblock Dinov2: Learning robust visual features without supervision.
\newblock 2023.

\bibitem[Paszke et~al.(2019)Paszke, Gross, Massa, Lerer, Bradbury, Chanan, Killeen, Lin, Gimelshein, Antiga, et~al.]{paszke2019pytorch}
Adam Paszke, Sam Gross, Francisco Massa, Adam Lerer, James Bradbury, Gregory Chanan, Trevor Killeen, Zeming Lin, Natalia Gimelshein, Luca Antiga, et~al.
\newblock Pytorch: An imperative style, high-performance deep learning library.
\newblock 2019.

\bibitem[Piccinelli et~al.(2024)Piccinelli, Yang, Sakaridis, Segu, Li, Van~Gool, and Yu]{piccinelli2024unidepth}
Luigi Piccinelli, Yung-Hsu Yang, Christos Sakaridis, Mattia Segu, Siyuan Li, Luc Van~Gool, and Fisher Yu.
\newblock Unidepth: Universal monocular metric depth estimation.
\newblock In \emph{CVPR}, 2024.

\bibitem[Ranftl et~al.(2020)Ranftl, Lasinger, Hafner, Schindler, and Koltun]{Ranftl2020TPAMI}
Ren{\'e} Ranftl, Katrin Lasinger, David Hafner, Konrad Schindler, and Vladlen Koltun.
\newblock Towards robust monocular depth estimation: Mixing datasets for zero-shot cross-dataset transfer.
\newblock \emph{IEEE TPAMI}, 2020.

\bibitem[Ranftl et~al.(2021)Ranftl, Bochkovskiy, and Koltun]{ranftl2021vision}
Ren{\'e} Ranftl, Alexey Bochkovskiy, and Vladlen Koltun.
\newblock Vision transformers for dense prediction.
\newblock In \emph{ICCV}, 2021.

\bibitem[Rombach et~al.(2022)Rombach, Blattmann, Lorenz, Esser, and Ommer]{rombach2022high}
Robin Rombach, Andreas Blattmann, Dominik Lorenz, Patrick Esser, and Bj{\"o}rn Ommer.
\newblock High-resolution image synthesis with latent diffusion models.
\newblock In \emph{CVPR}, 2022.

\bibitem[Ronneberger et~al.(2015)Ronneberger, Fischer, and Brox]{ronneberger2015u}
Olaf Ronneberger, Philipp Fischer, and Thomas Brox.
\newblock U-net: Convolutional networks for biomedical image segmentation.
\newblock Springer, 2015.

\bibitem[Sch{\"o}nberger et~al.(2016)Sch{\"o}nberger, Zheng, Frahm, and Pollefeys]{schonberger2016pixelwise}
Johannes~L Sch{\"o}nberger, Enliang Zheng, Jan-Michael Frahm, and Marc Pollefeys.
\newblock Pixelwise view selection for unstructured multi-view stereo.
\newblock In \emph{ECCV}, 2016.

\bibitem[Shi et~al.(2024)Shi, Wang, Ye, Long, Li, and Yang]{shi2023mvdream}
Yichun Shi, Peng Wang, Jianglong Ye, Mai Long, Kejie Li, and Xiao Yang.
\newblock Mvdream: Multi-view diffusion for 3d generation.
\newblock In \emph{ICLR}, 2024.

\bibitem[Suhail et~al.(2022)Suhail, Esteves, Sigal, and Makadia]{suhail2022generalizable}
Mohammed Suhail, Carlos Esteves, Leonid Sigal, and Ameesh Makadia.
\newblock Generalizable patch-based neural rendering.
\newblock In \emph{ECCV}. Springer, 2022.

\bibitem[Sun et~al.(2021)Sun, Xie, Chen, Zhou, and Bao]{sun2021neuralrecon}
Jiaming Sun, Yiming Xie, Linghao Chen, Xiaowei Zhou, and Hujun Bao.
\newblock Neuralrecon: Real-time coherent 3d reconstruction from monocular video.
\newblock In \emph{CVPR}, 2021.

\bibitem[Szymanowicz et~al.(2024)Szymanowicz, Rupprecht, and Vedaldi]{Szymanowicz2024CVPR}
Stanislaw Szymanowicz, Christian Rupprecht, and Andrea Vedaldi.
\newblock Splatter image: Ultra-fast single-view 3d reconstruction.
\newblock In \emph{CVPR}, 2024.

\bibitem[Szymanowicz et~al.(2025)Szymanowicz, Insafutdinov, Zheng, Campbell, Henriques, Rupprecht, and Vedaldi]{szymanowicz2024flash3d}
Stanislaw Szymanowicz, Eldar Insafutdinov, Chuanxia Zheng, Dylan Campbell, Jo{\~a}o~F Henriques, Christian Rupprecht, and Andrea Vedaldi.
\newblock Flash3d: Feed-forward generalisable 3d scene reconstruction from a single image.
\newblock In \emph{3DV}, 2025.

\bibitem[Tang and Tan(2018)]{tang2018ba}
Chengzhou Tang and Ping Tan.
\newblock Ba-net: Dense bundle adjustment networks.
\newblock In \emph{ICLR}, 2018.

\bibitem[Tang et~al.(2024)Tang, Chen, Chen, Wang, Zeng, and Liu]{tang2024lgm}
Jiaxiang Tang, Zhaoxi Chen, Xiaokang Chen, Tengfei Wang, Gang Zeng, and Ziwei Liu.
\newblock Lgm: Large multi-view gaussian model for high-resolution 3d content creation.
\newblock 2024.

\bibitem[Teed and Deng(2019)]{teed2019deepv2d}
Zachary Teed and Jia Deng.
\newblock Deepv2d: Video to depth with differentiable structure from motion.
\newblock In \emph{ICLR}, 2019.

\bibitem[Turkulainen et~al.(2025)Turkulainen, Ren, Melekhov, Seiskari, Rahtu, and Kannala]{turkulainen2024dn}
Matias Turkulainen, Xuqian Ren, Iaroslav Melekhov, Otto Seiskari, Esa Rahtu, and Juho Kannala.
\newblock Dn-splatter: Depth and normal priors for gaussian splatting and meshing.
\newblock In \emph{WACV}, 2025.

\bibitem[Ummenhofer et~al.(2017)Ummenhofer, Zhou, Uhrig, Mayer, Ilg, Dosovitskiy, and Brox]{ummenhofer2017demon}
Benjamin Ummenhofer, Huizhong Zhou, Jonas Uhrig, Nikolaus Mayer, Eddy Ilg, Alexey Dosovitskiy, and Thomas Brox.
\newblock Demon: Depth and motion network for learning monocular stereo.
\newblock In \emph{CVPR}, 2017.

\bibitem[Voleti et~al.(2024)Voleti, Yao, Boss, Letts, Pankratz, Tochilkin, Laforte, Rombach, and Jampani]{voleti2024sv3d}
Vikram Voleti, Chun-Han Yao, Mark Boss, Adam Letts, David Pankratz, Dmitry Tochilkin, Christian Laforte, Robin Rombach, and Varun Jampani.
\newblock Sv3d: Novel multi-view synthesis and 3d generation from a single image using latent video diffusion.
\newblock 2024.

\bibitem[Wang et~al.(2021)Wang, Galliani, Vogel, Speciale, and Pollefeys]{wang2021patchmatchnet}
Fangjinhua Wang, Silvano Galliani, Christoph Vogel, Pablo Speciale, and Marc Pollefeys.
\newblock Patchmatchnet: Learned multi-view patchmatch stereo.
\newblock In \emph{CVPR}, 2021.

\bibitem[Wang et~al.(2022)Wang, Galliani, Vogel, and Pollefeys]{wang2021itermvs}
Fangjinhua Wang, Silvano Galliani, Christoph Vogel, and Marc Pollefeys.
\newblock Itermvs: Iterative probability estimation for efficient multi-view stereo.
\newblock In \emph{CVPR}, 2022.

\bibitem[Wang et~al.(2020)Wang, Zhu, Wang, Hu, Qiu, Wang, Hu, Kapoor, and Scherer]{wang2020tartanair}
Wenshan Wang, Delong Zhu, Xiangwei Wang, Yaoyu Hu, Yuheng Qiu, Chen Wang, Yafei Hu, Ashish Kapoor, and Sebastian Scherer.
\newblock Tartanair: A dataset to push the limits of visual slam.
\newblock In \emph{2020 IEEE/RSJ International Conference on Intelligent Robots and Systems (IROS)}, pages 4909--4916. IEEE, 2020.

\bibitem[Wang et~al.(2023)Wang, Shi, Li, Huang, Cao, Zhang, Xian, and Lin]{wang2023neural}
Yiran Wang, Min Shi, Jiaqi Li, Zihao Huang, Zhiguo Cao, Jianming Zhang, Ke Xian, and Guosheng Lin.
\newblock Neural video depth stabilizer.
\newblock In \emph{ICCV}, 2023.

\bibitem[Wang et~al.(2024)Wang, Huang, Chen, and Lee]{wang2024freesplat}
Yunsong Wang, Tianxin Huang, Hanlin Chen, and Gim~Hee Lee.
\newblock Freesplat: Generalizable 3d gaussian splatting towards free view synthesis of indoor scenes.
\newblock In \emph{NeurIPS}, 2024.

\bibitem[Weng et~al.(2023)Weng, Yang, Wang, Li, Zhang, Chen, and Zhang]{weng2023consistent123}
Haohan Weng, Tianyu Yang, Jianan Wang, Yu Li, Tong Zhang, CL Chen, and Lei Zhang.
\newblock Consistent123: Improve consistency for one image to 3d object synthesis.
\newblock 2023.

\bibitem[Wewer et~al.(2024)Wewer, Raj, Ilg, Schiele, and Lenssen]{wewer2024latentsplat}
Christopher Wewer, Kevin Raj, Eddy Ilg, Bernt Schiele, and Jan~Eric Lenssen.
\newblock latentsplat: Autoencoding variational gaussians for fast generalizable 3d reconstruction.
\newblock In \emph{ECCV}, 2024.

\bibitem[Xu and Zhang(2020)]{xu2020aanet}
Haofei Xu and Juyong Zhang.
\newblock Aanet: Adaptive aggregation network for efficient stereo matching.
\newblock In \emph{CVPR}, 2020.

\bibitem[Xu et~al.(2022)Xu, Zhang, Cai, Rezatofighi, and Tao]{xu2022gmflow}
Haofei Xu, Jing Zhang, Jianfei Cai, Hamid Rezatofighi, and Dacheng Tao.
\newblock Gmflow: Learning optical flow via global matching.
\newblock In \emph{CVPR}, 2022.

\bibitem[Xu et~al.(2023)Xu, Zhang, Cai, Rezatofighi, Yu, Tao, and Geiger]{xu2023unifying}
Haofei Xu, Jing Zhang, Jianfei Cai, Hamid Rezatofighi, Fisher Yu, Dacheng Tao, and Andreas Geiger.
\newblock Unifying flow, stereo and depth estimation.
\newblock \emph{IEEE TPAMI}, 2023.

\bibitem[Xu et~al.(2024{\natexlab{a}})Xu, Chen, Chen, Sakaridis, Zhang, Pollefeys, Geiger, and Yu]{xu2023murf}
Haofei Xu, Anpei Chen, Yuedong Chen, Christos Sakaridis, Yulun Zhang, Marc Pollefeys, Andreas Geiger, and Fisher Yu.
\newblock Murf: Multi-baseline radiance fields.
\newblock In \emph{CVPR}, 2024{\natexlab{a}}.

\bibitem[Xu and Tao(2019)]{xu2019acmm}
Qingshan Xu and Wenbing Tao.
\newblock Multi-scale geometric consistency guided multi-view stereo.
\newblock In \emph{CVPR}, 2019.

\bibitem[Xu et~al.(2024{\natexlab{b}})Xu, Shi, Yifan, Chen, Yang, Peng, Shen, and Wetzstein]{xu2024grm}
Yinghao Xu, Zifan Shi, Wang Yifan, Hansheng Chen, Ceyuan Yang, Sida Peng, Yujun Shen, and Gordon Wetzstein.
\newblock Grm: Large gaussian reconstruction model for efficient 3d reconstruction and generation.
\newblock 2024{\natexlab{b}}.

\bibitem[Yang et~al.(2024{\natexlab{a}})Yang, Kang, Huang, Xu, Feng, and Zhao]{Yang2024CVPR}
Lihe Yang, Bingyi Kang, Zilong Huang, Xiaogang Xu, Jiashi Feng, and Hengshuang Zhao.
\newblock Depth anything: Unleashing the power of large-scale unlabeled data.
\newblock In \emph{CVPR}, 2024{\natexlab{a}}.

\bibitem[Yang et~al.(2024{\natexlab{b}})Yang, Kang, Huang, Zhao, Xu, Feng, and Zhao]{yang2024depth}
Lihe Yang, Bingyi Kang, Zilong Huang, Zhen Zhao, Xiaogang Xu, Jiashi Feng, and Hengshuang Zhao.
\newblock Depth anything v2.
\newblock 2024{\natexlab{b}}.

\bibitem[Yao et~al.(2018)Yao, Luo, Li, Fang, and Quan]{yao2018mvsnet}
Yao Yao, Zixin Luo, Shiwei Li, Tian Fang, and Long Quan.
\newblock Mvsnet: Depth inference for unstructured multi-view stereo.
\newblock In \emph{ECCV}, 2018.

\bibitem[Ye et~al.(2025)Ye, Liu, Xu, Li, Pollefeys, Yang, and Peng]{ye2024no}
Botao Ye, Sifei Liu, Haofei Xu, Xueting Li, Marc Pollefeys, Ming-Hsuan Yang, and Songyou Peng.
\newblock No pose, no problem: Surprisingly simple 3d gaussian splats from sparse unposed images.
\newblock In \emph{ICLR}, 2025.

\bibitem[Yin et~al.(2022)Yin, Zhang, Wang, Niklaus, Chen, Liu, and Shen]{yin2022towards}
Wei Yin, Jianming Zhang, Oliver Wang, Simon Niklaus, Simon Chen, Yifan Liu, and Chunhua Shen.
\newblock Towards accurate reconstruction of 3d scene shape from a single monocular image.
\newblock \emph{IEEE TPAMI}, 2022.

\bibitem[Yin et~al.(2023)Yin, Zhang, Chen, Cai, Yu, Wang, Chen, and Shen]{Yin2023CVPR}
Wei Yin, Chi Zhang, Hao Chen, Zhipeng Cai, Gang Yu, Kaixuan Wang, Xiaozhi Chen, and Chunhua Shen.
\newblock Metric3d: Towards zero-shot metric 3d prediction from a single image.
\newblock In \emph{CVPR}, 2023.

\bibitem[Yu et~al.(2021)Yu, Ye, Tancik, and Kanazawa]{yu2021pixelnerf}
Alex Yu, Vickie Ye, Matthew Tancik, and Angjoo Kanazawa.
\newblock pixelnerf: Neural radiance fields from one or few images.
\newblock In \emph{CVPR}, 2021.

\bibitem[Zhang et~al.(2025{\natexlab{a}})Zhang, Zou, Li, Yi, and Wang]{zhang2024transplat}
Chuanrui Zhang, Yingshuang Zou, Zhuoling Li, Minmin Yi, and Haoqian Wang.
\newblock Transplat: Generalizable 3d gaussian splatting from sparse multi-view images with transformers.
\newblock In \emph{AAAI}, 2025{\natexlab{a}}.

\bibitem[Zhang et~al.(2024{\natexlab{a}})Zhang, Bi, Tan, Xiangli, Zhao, Sunkavalli, and Xu]{zhang2024gs}
Kai Zhang, Sai Bi, Hao Tan, Yuanbo Xiangli, Nanxuan Zhao, Kalyan Sunkavalli, and Zexiang Xu.
\newblock Gs-lrm: Large reconstruction model for 3d gaussian splatting.
\newblock 2024{\natexlab{a}}.

\bibitem[Zhang et~al.(2018)Zhang, Isola, Efros, Shechtman, and Wang]{zhang2018unreasonable}
Richard Zhang, Phillip Isola, Alexei~A Efros, Eli Shechtman, and Oliver Wang.
\newblock The unreasonable effectiveness of deep features as a perceptual metric.
\newblock In \emph{CVPR}, 2018.

\bibitem[Zhang et~al.(2024{\natexlab{b}})Zhang, Fei, Liu, Song, and Duan]{zhang2024gaussian}
Shengjun Zhang, Xin Fei, Fangfu Liu, Haixu Song, and Yueqi Duan.
\newblock Gaussian graph network: Learning efficient and generalizable gaussian representations from multi-view images.
\newblock In \emph{NeurIPS}, 2024{\natexlab{b}}.

\bibitem[Zhang et~al.(2025{\natexlab{b}})Zhang, Wang, Xu, Xue, Rupprecht, Zhou, Shen, and Wetzstein]{zhang2025flare}
Shangzhan Zhang, Jianyuan Wang, Yinghao Xu, Nan Xue, Christian Rupprecht, Xiaowei Zhou, Yujun Shen, and Gordon Wetzstein.
\newblock Flare: Feed-forward geometry, appearance and camera estimation from uncalibrated sparse views.
\newblock In \emph{CVPR}, 2025{\natexlab{b}}.

\bibitem[Zheng and Vedaldi(2024)]{zheng2023free3d}
Chuanxia Zheng and Andrea Vedaldi.
\newblock Free3d: Consistent novel view synthesis without 3d representation.
\newblock In \emph{CVPR}, 2024.

\bibitem[Zhou et~al.(2018)Zhou, Tucker, Flynn, Fyffe, and Snavely]{zhou2018stereo}
Tinghui Zhou, Richard Tucker, John Flynn, Graham Fyffe, and Noah Snavely.
\newblock Stereo magnification: learning view synthesis using multiplane images.
\newblock \emph{ACM TOG}, 2018.

\end{thebibliography}
}

\clearpage
\maketitlesupplementary

\counterwithin{figure}{section}
\counterwithin{table}{section}
\renewcommand\thesection{\Alph{section}}
\renewcommand\thetable{\thesection.\arabic{table}}
\renewcommand\thefigure{\thesection.\arabic{figure}}
\setcounter{section}{0}

\section{Depth Pre-Training for Gaussian Splatting}
\label{sec:depth2gs}

In the main paper, we have shown that better depth architectures lead to improved view synthesis with Gaussian splatting. We further study the effect of different weight initializations for the depth model. Specifically, we compare three variants: 1) only initializing the monocular feature with Depth Anything V2~\citep{yang2024depth}; 2) initializing the monocular feature with Depth Anything V2~\citep{yang2024depth} and the multi-view feature with UniMatch~\citep{xu2023unifying}; 3) initializing the full depth model by pre-training the depth model on TartanAir~\cite{wang2020tartanair}. We can observe from \cref{tab:depth2gs} that better depth initialization also leads to improved view synthesis results.

\begin{table}[!h]
\begin{center}
\small
\caption{\textbf{Depth pre-training for Gaussian splatting}. We compare different weight initializations for the depth model when training our full DepthSplat model for view synthesis. Compared to 1) only initializing the monocular feature (mono features) and 2) initializing the monocular and multi-view features (mono \& mv features), our pre-trained full depth model (full depth model) achieves the best view synthesis results.}

\begin{tabular}{ccccccccccccccccccccccc}
\toprule

Initialization & PSNR $\uparrow$ & SSIM $\uparrow$ & LPIPS $\downarrow$ \\
    
    \toprule

    mono features & 26.59 & 0.874 & 0.1256 \\
    mono \& mv features & 26.76 & 0.877 & 0.1234 \\
    full depth model & \textbf{26.81} & \textbf{0.878} & \textbf{0.1225} \\

\bottomrule
\end{tabular}
\label{tab:depth2gs}

\end{center}
\end{table}

\section{More Visual Results}
\label{sec:supp_vis}

\subsection{Unsupervised Depth Pre-Training with Gaussian Splatting}

\begin{figure*}[!t]
    \centering
    \includegraphics[width=\linewidth]{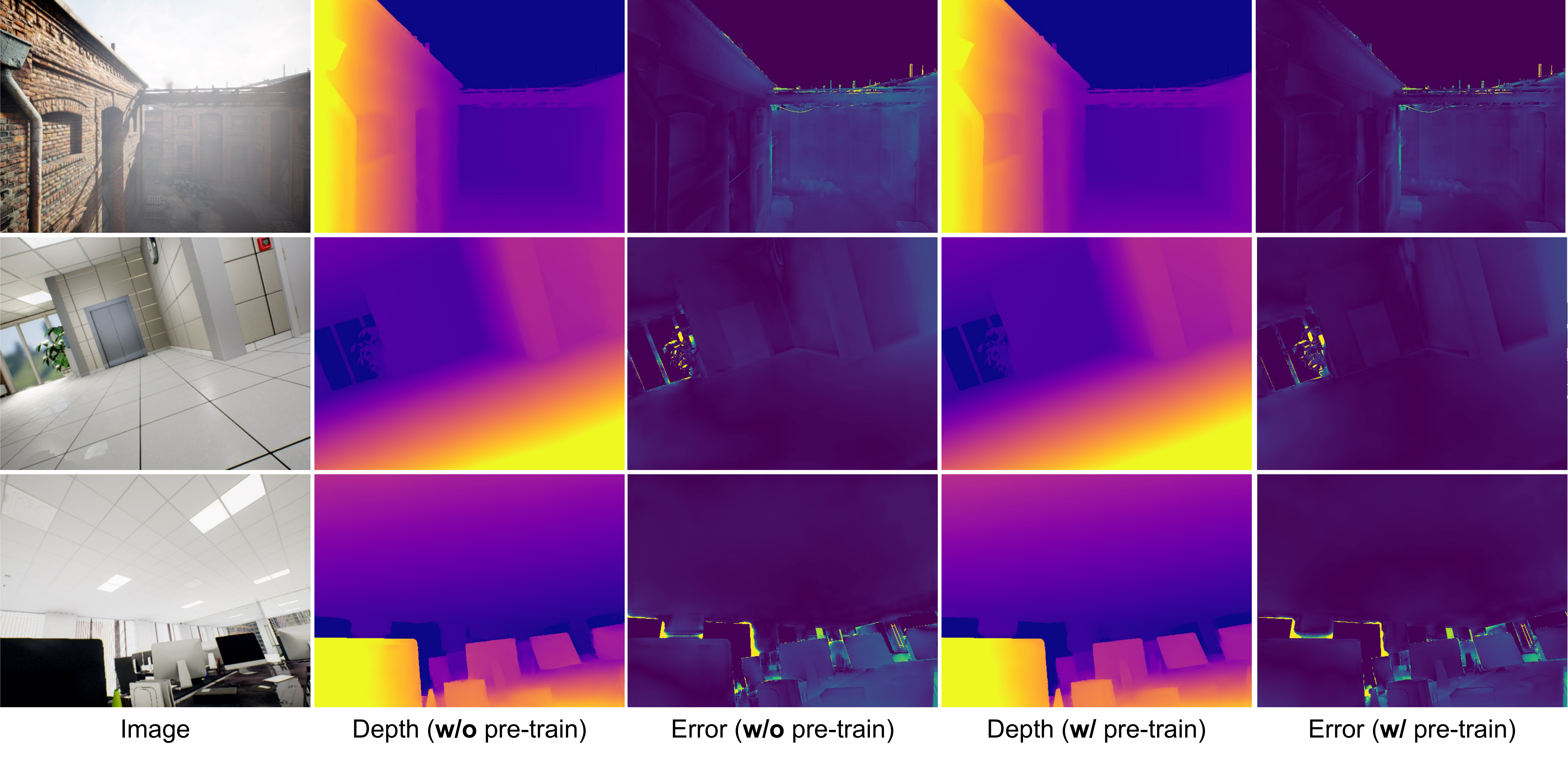}
    \caption{\textbf{Effect of unsupervised depth pre-training}. We observe that the unsupervised pre-training leads to improved performance for texture-less regions.
    }
    \label{fig:depth_pretrain}
\end{figure*}

In \cref{fig:depth_pretrain}, we show the comparisons of depth estimation results with and without Gaussian splatting pre-training. We observe that pre-training leads to better results in texture-less regions. We hypothesize that pre-training provides regularizations (as also observed in ~\cite{erhan2010does}) to the challenging scenarios and accordingly leads to improved performance.

\subsection{Cross-Dataset Generalization}

\begin{figure*}[!t]
    \centering
    \includegraphics[width=\linewidth]{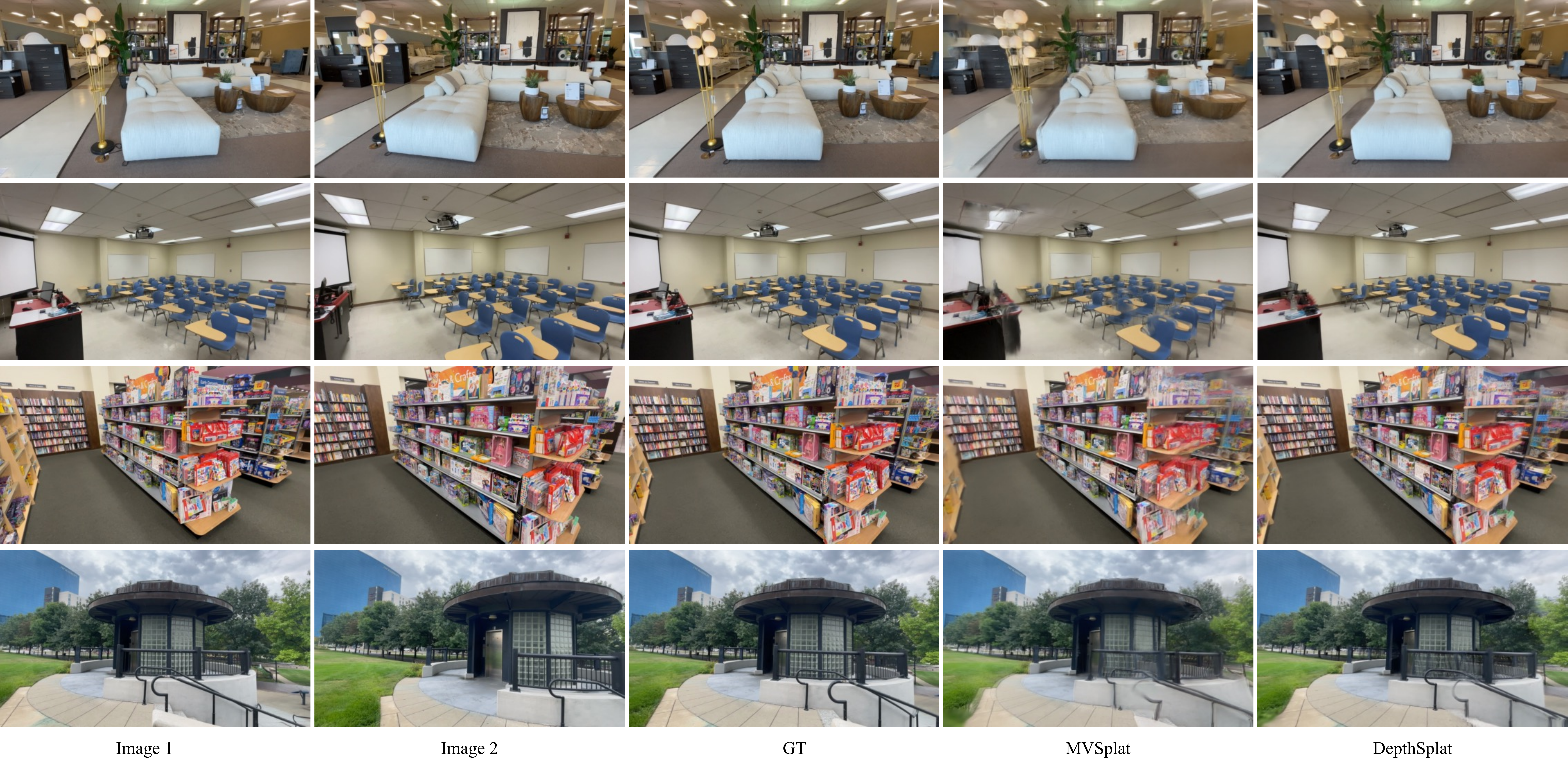}
    \caption{\textbf{Generalization from RealEstate10K to DL3DV}.
    }
    \label{fig:re10k_to_dl3dv}
\end{figure*}

In \cref{fig:re10k_to_dl3dv}, we show the cross-dataset generalization results on the DL3DV~\cite{ling2024dl3dv} dataset, which are obtained with the RealEstate10K~\cite{zhou2018stereo} pre-trained models. Our DepthSplat generalizes more robustly than MVSplat~\cite{Chen2024ARXIV} on unseen scenes.

\subsection{Visual Comparisons on DL3DV}

\begin{figure*}[!t]
    \centering
    \includegraphics[width=\linewidth]{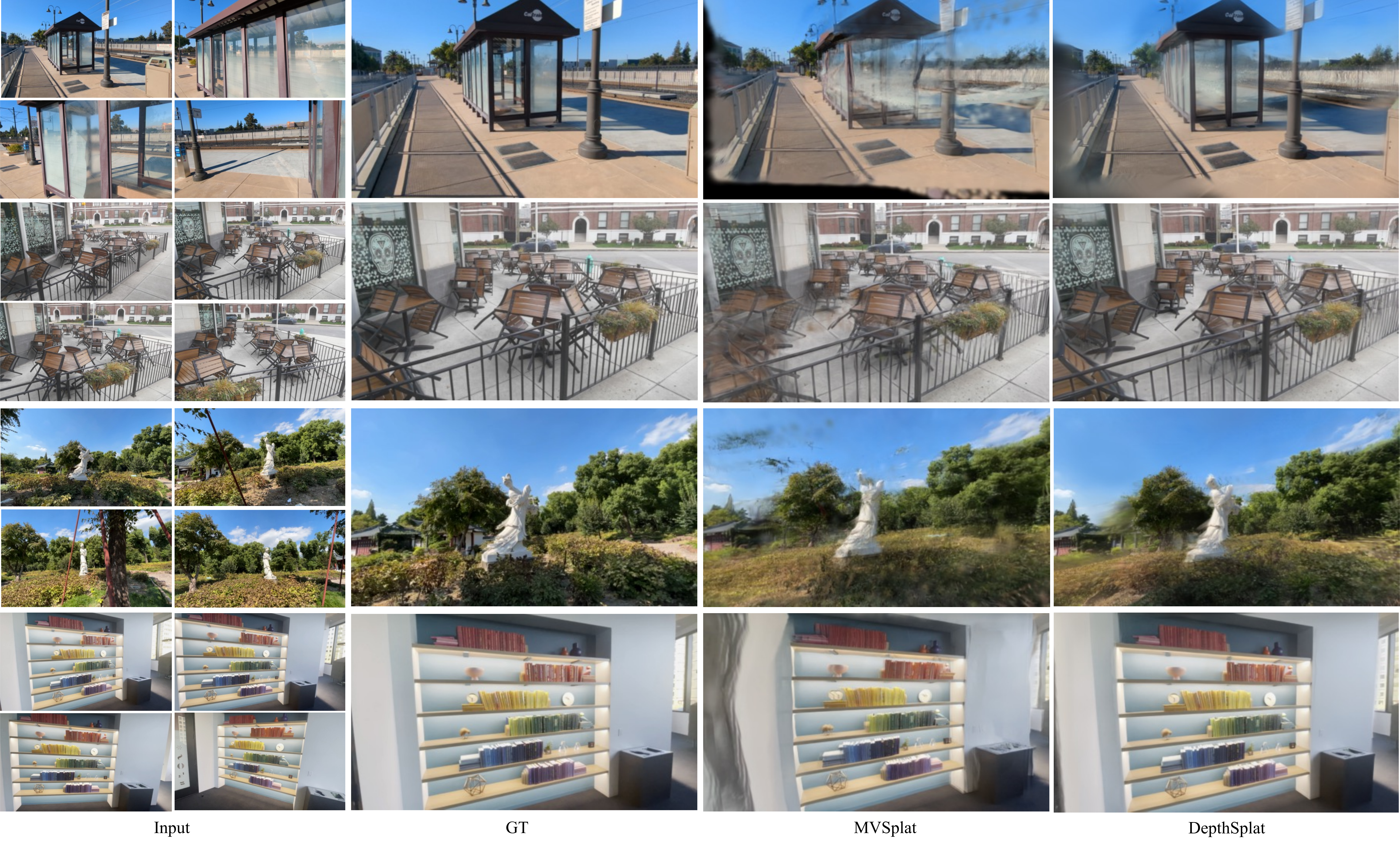}
    \vspace{-5mm}
    \caption{\textbf{View synthesis from 4 input views on DL3DV}.
    }
    \label{fig:dl3dv_render_4view}
\end{figure*}

In \cref{fig:dl3dv_render_4view}, we compare the visual synthesis results from 4 input views with MVSplat~\cite{Chen2024ARXIV}. Our DepthSplat better preserves the scene structures.

\subsection{High-Resolution Results}

\begin{figure*}[!t]
    \centering
    \includegraphics[width=\linewidth]{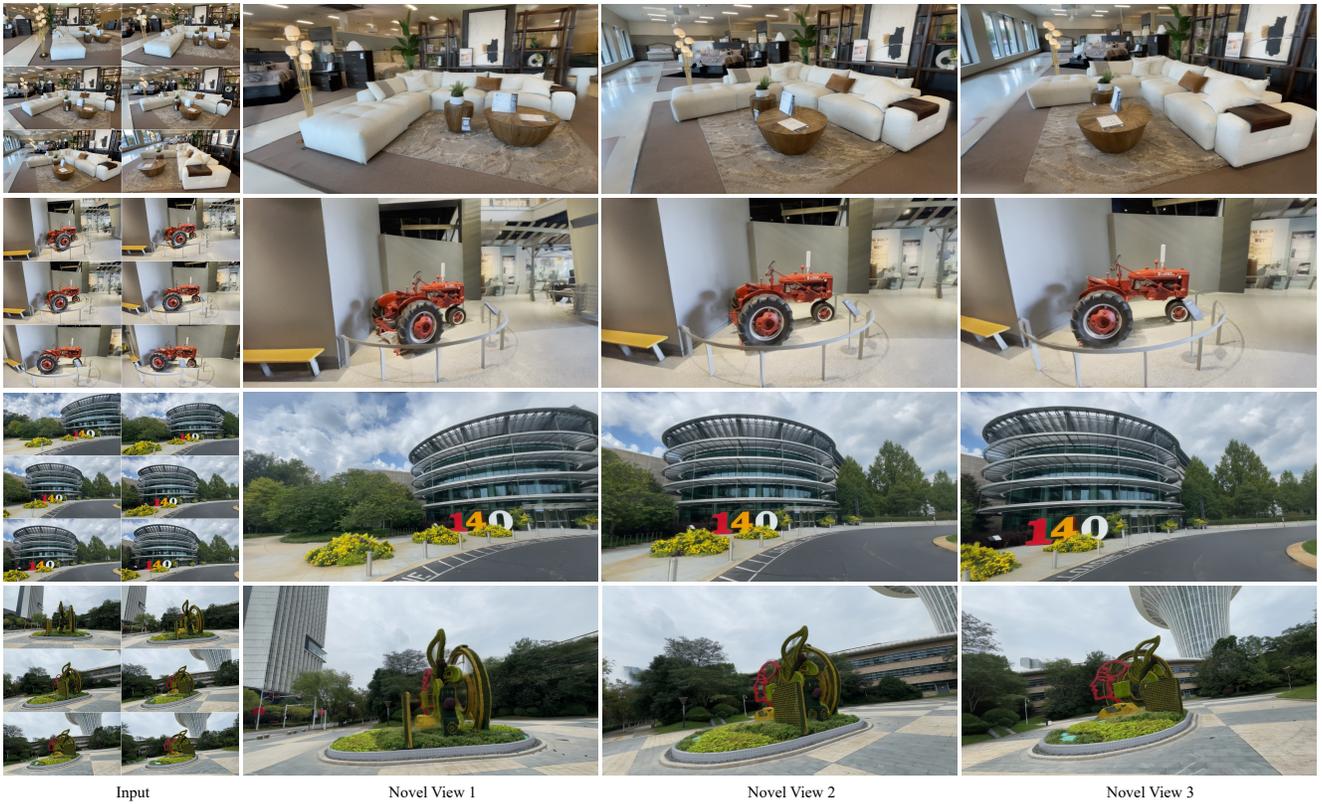}
    \vspace{-5mm}
    \caption{\textbf{View synthesis at $512 \times 960$ resolutions from 6 input views}.
    }
    \label{fig:dl3dv_render_512x960}
    \vspace{-4mm}
\end{figure*}

\begin{figure*}[!t]
    \centering
    \includegraphics[width=\linewidth]{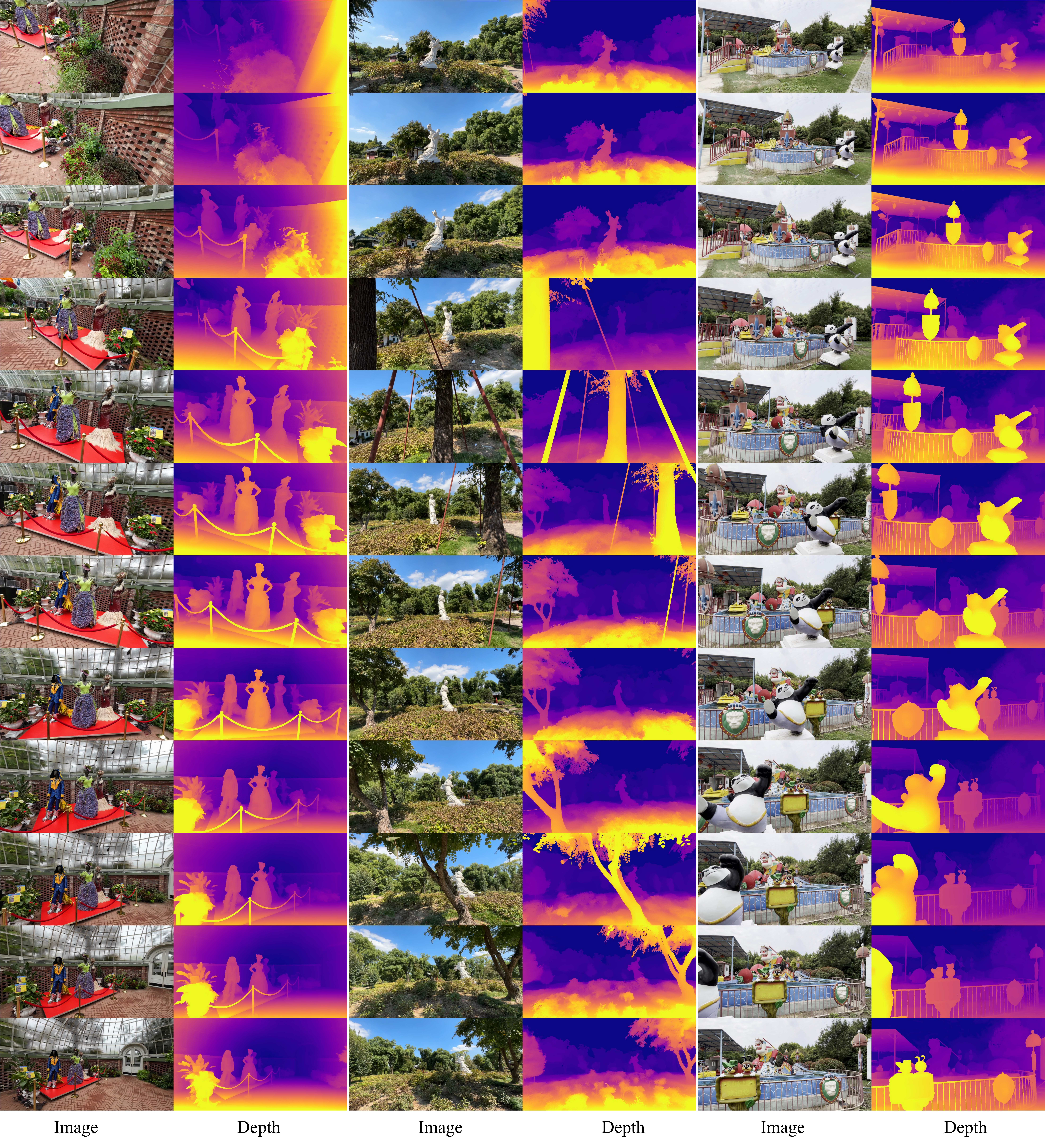}
    \caption{\textbf{Depth predictions on DL3DV with 12 input views}. The image resolutions are $512 \times 960$.
    }
    \label{fig:dl3dv_mvdepth}
\end{figure*}

In \cref{fig:dl3dv_render_512x960} and \cref{fig:dl3dv_mvdepth}, we show the view synthesis and multi-view depth estimation results at $512 \times 960$ resolutions. Note that MVSplat~\cite{Chen2024ARXIV} runs out-of-memory on such high resolutions with 6 or 12 input views, while our DepthSplat significantly improves the efficiency with two technical components. First, our lowest feature resolution is $1/8$ of the image resolution, while MVSplat uses $1/4$. Second, we use local cross-view attentions unlike the pair-wise global attentions in MVSplat.

\section{More Implementation Details}
\label{sec:supp_details}

We provide more implementation details on the high-resolution experiments. For high-resolution experiments, we choose our small model which contains a ViT-S monocular branch and a single-scale multi-view branch. We first train our model on RealEstate10K with two input views at $256 \times 448$ resolutions for 150K iterations, where the total batch size is 256. We fine-tune the model on the mixed RealEstate10K and DL3DV datasets at $448 \times 768$ resolutions for 200K iterations with a total batch size of 64, where the number of input views is randomly sampled from 4 to 10. We use this model to predict results on high resolutions (\eg, $512 \times 960$) and different numbers of input views (\eg, 6 and 12), as we show in our project page: \href{https://haofeixu.github.io/depthsplat/}{haofeixu.github.io/depthsplat}.

\end{document}